\documentclass{article}
 \pdfoutput=1  
\usepackage{amssymb}

\usepackage{scalefnt,letltxmacro}
\LetLtxMacro{\oldtextsc}{\textsc}
\renewcommand{\textsc}[1]{\oldtextsc{\scalefont{1.10}#1}}
\usepackage{mathtools}
\usepackage{amsfonts}
\usepackage{amsmath}
\usepackage{amsthm}

\theoremstyle{definition}

\usepackage{empheq}

\usepackage[english]{babel}
\usepackage[parfill]{parskip}
\usepackage{afterpage}
\usepackage{enumitem}
\usepackage{framed}
\usepackage{setspace}

\usepackage{ragged2e}

\newcounter{parcount}
\newcommand\p{%
    \stepcounter{parcount}%
    \leavevmode\marginpar[\hfill\parnum]{\parnum}%
}

\usepackage{rotating}

\usepackage[usenames,dvipsnames]{xcolor}
\definecolor{shadecolor}{gray}{0.9}

\usepackage{graphicx}
\usepackage[labelfont=bf]{caption}
\usepackage[format=hang]{subcaption}
\usepackage{wrapfig}

\usepackage{booktabs}
\usepackage{array}
\usepackage{multirow}

\usepackage[colorlinks,linktoc=all]{hyperref}
\usepackage[all]{hypcap}
\hypersetup{citecolor=MidnightBlue}
\hypersetup{linkcolor=MidnightBlue}
\hypersetup{urlcolor=MidnightBlue}

\usepackage[acronym,smallcaps,nowarn]{glossaries}
\glsdisablehyper
\makeglossaries

\usepackage{listings}

\definecolor{strings}{rgb}{.624,.251,.259}
\definecolor{keywords}{rgb}{.224,.451,.686}
\definecolor{comment}{rgb}{.322,.451,.322}

\lstdefinelanguage{python}{
  morekeywords={from, import, as, for, in, while, def, return, =, +,
  -, /, *, lambda},
  keywords=[2]{build_toy_dataset, neural_network, __init__},
  keywords=[3]{Normal, Bernoulli, Beta, Categorical, Dirichlet,
  Exponential, MultivariateNormalFull, RandomVariable, Distribution,
  DirichletProcess, Empirical, PointMass, Gamma,
  MAP, Inference, KLqp, HMC, SGLD, KLpq,
  VariationalInference, MonteCarlo, ConjugateInference, GANInference,
  rnn_cell, dirichlet_process, cond, body,
  evaluate, ppc, copy, dot, get_session},
  morecomment=[l]{\#},
  morecomment=[s]{"""}{"""},
  morestring=[b]',
  morestring=[b]",
  alsoletter={<>=-+/*},
  sensitive=true
}

\renewcommand{\texttt}[1]{\lstinline[basicstyle=\fontsize{8pt}{8.25pt}\selectfont\ttfamily]{#1}}

\usepackage[nameinlink]{cleveref}
\crefname{section}{\S}{\S\S}
\Crefname{section}{\S}{\S\S}
\creflabelformat{equation}{#2\textup{#1}#3}

\usepackage[colorinlistoftodos,
            textsize=tiny,
            linecolor=red!30,
            bordercolor=red!30,
            backgroundcolor=red!10]{todonotes}

\usepackage{xspace}
\usepackage[most]{tcolorbox}

\newtcolorbox{incomplete}{skin=enhancedmiddle jigsaw,breakable,
  parbox=false,boxrule=0mm,leftrule=2mm,boxsep=0mm,
  arc=0mm,outer arc=0mm,left=3mm,right=3mm,top=1mm,bottom=1mm,
  toptitle=1mm,bottomtitle=1mm,oversize,colback=yellow!5!white,
  colframe=yellow}

\newcommand{\q}[1]{\red{{\sf Q | #1}}}

\newacronym{RMSE}{rmse}{root mean squared error}
\newacronym{MAE}{mae}{mean absolute error}
\newacronym{NLPD}{nlpd}{negative log predictive density}
\newacronym{MRANK}{m-rank}{average model rank}
\newacronym{FVAR}{f-var}{forecast variance}

\newacronym{GP}{gp}{Gaussian process}
\newacronym{GGP}{ggp}{grouped Gaussian process}
\newacronym{INLA}{inla}{integrated nested Laplace approximation}

\newacronym{ELBO}{elbo}{evidence lower bound}
\newacronym{ELL}{ell}{expected log likelihood}
\newacronym{KL}{kl}{Kullback-Leibler}

\newacronym{PV}{pv}{photovoltaic}

\newcommand{\dataset}{\mathcal{D}}

\newacronym{MAP}{map}{maximum a posteriori}
\newacronym{MLE}{mle}{maximum likelihood estimation}

\newacronym{VAE}{vae}{variational autoencoder}

\newacronym{MC}{mc}{Monte Carlo}
\newacronym{MCMC}{mcmc}{Markov chain Monte Carlo}

\newacronym{VI}{vi}{variational inference}

\newacronym{MLP}{mlp}{multilayer perceptron}
\newacronym{RELU}{ReLU}{rectified linear unit}

\newacronym{CNN}{cnn}{convolutional neural network}
\newacronym{GCN}{gcn}{graph convolutional network}
\newacronym{CHEBNET}{chebnet}{Chebyshev network}
\newacronym{GRAPHSAGE}{graphsage}{sample-and-aggregate}
\newacronym{GAT}{gat}{graph attention network}
\newacronym{MCD}{mcd}{Monte Carlo dropout}
\newacronym{NNMF}{nnmf}{neural network matrix factorization}
\newacronym{PMF}{pmf}{probabilistic matrix factorization}
\newacronym{MONET}{monet}{mixture models network} 

\newacronym{ARD}{ard}{automatic relevance determination}

\newacronym{SVD}{svd}{singular value decomposition}

\newacronym{CF}{cf}{collaborative filtering}
\newacronym{LP}{lp}{link prediction}

\usepackage[preprint,nonatbib]{neurips_2019}
\newcommand{\citep}{\cite}

\usepackage[utf8]{inputenc} % allow utf-8 input
\usepackage[T1]{fontenc}    % use 8-bit T1 fonts
\usepackage{url}            % simple URL typesetting
\usepackage{booktabs}       % professional-quality tables
\usepackage{amsfonts}       % blackboard math symbols
\usepackage{nicefrac}       % compact symbols for 1/2, etc.
\usepackage{microtype}      % microtypography

\newcommand{\cond}{\,|\,}
\newcommand{\ie}{i.e.\xspace}
\newcommand{\eg}{e.g.\xspace}

\newcommand{\apriori}{\emph{a priori}\xspace}

\newcommand{\mat}[1]{\mathbf{#1}}
\renewcommand{\vec}[1]{ \mathbf{#1} } % math bold
\newcommand{\vecS}[1]{\boldsymbol{ #1 }  } % this for boldsymbols

\newcommand{\F}{\mat{F}}

\newcommand{\I}{\mat{I}}
\newcommand{\K}{\mat{K}}

\newcommand{\X}{\mat{X}}
\newcommand{\W}{\mat{W}}
\newcommand{\Y}{\mat{Y}}
\newcommand{\Z}{\mat{Z}}

\newcommand{\calL}{\mathcal{L}}

\newcommand{\calO}{\mathcal{O}}

\newcommand{\vecmu}{\vecS{\mu}}

\newcommand{\setR}{\mathbb{R}}

\newcommand{\kron}{\otimes}

\newcommand{\expectation}[2]{ \mathbb{E}_{#1}{\left[#2\right]} }

\newcommand{\Normal}{\mathcal{N}}

\newcommand{\trace}{\mbox{ \rm tr }}
\renewcommand{\det}[1]{\left\lvert#1\right\rvert}
\newcommand{\defeq}{\stackrel{\text{\tiny def}}{=}}

\newcommand{\bigO}{\calO}

\newcommand{\name}[1]{{\sc #1}\xspace}

\newcommand{\mtg}{\name{mtg}}

\newcommand{\gprn}{\name{gprn}}

\newcommand{\lcm}{\name{lcm}}
\newcommand{\adam}{\name{adam}}

\newcommand{\elbo}{\calL_{\text{elbo}}}
\newcommand{\df}{\text{d}\f}

\newcommand{\ellterm}{\calL_{\text{ell}}}

\newcommand{\enterm}{\calL_{\text{ent}}}
\newcommand{\crossterm}{\calL_{\text{cross}}}

\newcommand{\nlpd}{\name{nlpd}}

\newcommand{\rmse}{\name{rmse}}

\newcommand{\fvar}{\name{f-var}}
\newcommand{\mrank}{\name{m-rank}}
\newcommand{\mae}{\name{mae}}

\newcommand{\gp}{\mathcal{GP}}
\newcommand{\meanfun}{\mu}
\newcommand{\kernel}{\kappa}

\newcommand{\hyperparam}{\vecS{\theta}}
\newcommand{\likeparam}{\vecS{\phi}}

\newcommand{\pcond}[2]{p(#1  | #2 )}

\newcommand{\bs}{\boldsymbol}
\newcommand{\llambda}{\bs{\lambda}}

\newcommand{\s}{S} % Number of samples
\newcommand{\n}{N} % Number of observatios
\newcommand{\m}{M} % number of inducing points
\renewcommand{\p}{P} % Number of output variables
\renewcommand{\q}{Q} % Number of latent processes
\renewcommand{\d}{D} % Input dimension
\renewcommand{\k}{K} % Number of mixture components
\newcommand{\dimh}{H} % Input dimension of task-specific features

\newcommand{\vecy}{\vec{y}} % vector of output values 
\newcommand{\y}{\Y} % matrix of all output values 
\newcommand{\yn}{\vecy_{(n)}} % vector of output values at n
\newcommand{\xn}{\vec{x}_{(n)}}
\newcommand{\f}{\F} % matrix of all latent function values 
\newcommand{\vecf}{\vec{f}} %vector of latent function values 
\newcommand{\fn}{\vecf_{(n)}} % vector of all latent function values at n with dot
\newcommand{\fnnodot}{\vecf_{(n)}} % vector of all latent function values at n without dot
\renewcommand{\u}{\vec{u}} % vector of all inducing points
\newcommand{\x}{\vec{x}}
\newcommand{\xprime}{\x^\prime}

\newcommand{\z}{\mat{Z}}
\renewcommand{\X}{\mat{X}}
\newcommand{\vecz}{\vec{z}}

\newcommand{\lags}{\boldsymbol{l}} % lag feature vector
\newcommand{\h}{\vec{h}} % spatial feature vector 
\newcommand{\hprime}{\h^\prime}
\newcommand{\bR}{R} % Number of blocks in GGP
\newcommand{\br}{r} % block index, r=1 to R
\newcommand{\Qr}{Q_r} % number of latent processes in block r
\newcommand{\Qg}{Q_g} % Number of node processes

\newcommand{\fr}{\f_{\br}} % matrix of latent function values, j in block r
\newcommand{\Wn}{\mat{W}_{(n)}} %W(x) matrix weight functions
\newcommand{\g}{\mat{G}}
\newcommand{\vecg}{\vec{g}}
\newcommand{\gj}{\vec{g}_j}  % updated AAAI (AD)
\newcommand{\gn}{\vec{g}_{(n)}} %g(_n) vector node functions
\newcommand{\Krxxhh}{\K_{f f}^\br} %replacement AAAI (AD)
\newcommand{\Krxx}{\K_{\x\x}^\br} %   

\newcommand{\Krhh}{\K_{\h\h}^\br} %spatial link gram matrix
\newcommand{\Krhhinv}{(\K_{\h\h}^\br)^{-1}} % spatial link mat inv
\newcommand{\Krzz}{\K_{\vecz\vecz}^\br} %\newcommand{\Kzz}{\kernel(\Zr, \Zr)}
\newcommand{\Krzzinv}{(\K_{\vecz\vecz}^\br)^{-1}} %\newcommand{\Kzz}{\kernel(\Zr, \Zr)}
\newcommand{\Krxz}{\K_{\x\vecz}^\br} %\newcommand{\Kxz}{\kernel(\X,\Z_r)}
\newcommand{\Krzzhh}{\K_{u u}^\br} %\newcommand{\Kzz}{\kernel(\Zr, \Zr)}
\newcommand{\Krzzhhinv}{(\K_{u u}^\br)^{-1}} %\newcommand{\Kzz}{\kernel(\Zr, \Zr)}
\newcommand{\Krxzh}{\K_{f u}^\br} %\newcommand{\Kxz}{\kernel(\X,\Z_r)}
\newcommand{\Krzxh}{\K_{u f}^\br} %\newcommand{\Kzx}{\kernel(\Z_r, \X)}
\newcommand{\postcovkronb}{\mat{S}_{k\br b}}
\newcommand{\postcovkronw}{\mat{S}_{k\br w}}

\newcommand{\ur}{\vec{u}_{ \br} } % inducing points for latent block r
\newcommand{\Zr}{\z_\br} % inducing inputs for latent block r
\newcommand{\priormeanr}{\tilde{\vecS{\mu}}_\br} % prior mean
\newcommand{\priorcovr}{\widetilde{\K}_\br}
\newcommand{\Ar}{\mat{A}_\br}
\newcommand{\Hmat}{\mat{H}}
\newcommand{\Hrmat}{\mat{H}_\br}
\newcommand{\priorcovrn}{\widetilde{\K}_{\br (n)}}
\newcommand{\Arn}{\mat{A}_{\br (n)}}

\newcommand{\krxh}{\kernel_{\br}(f_j(\x), f_{j'}(\xprime) )}

\newcommand{\krxx}{\kernel_{\br}(\x, \x')}
\newcommand{\krxxn}{\kernel_{\br}(\x_{(n)}, \x_{(n)}}
\newcommand{\krxzn}{\kernel_{\br}(\x_{(n)},\Z_r)}
\newcommand{\krzxn}{\kernel_{\br}(\Z_r, \x_{(n)})}

\newcommand{\krhh}{\kernel_{\br}( \h_j, \h_{j'} )}
\newcommand{\ksub}[1]{\kernel_{#1} }

\newcommand{\kprop}{\pi_k}
\newcommand{\qukr}{q_k(\ur | \llambda_{k \br})}
\newcommand{\varparamkr}{\llambda_{k \br}}

\newcommand{\xstar}{\x_{\star}}
\newcommand{\fstar}{\vecf_{\star}}

\newcommand{\ystar}{\vecy_{\star}}

\renewcommand{\K}{\mat{K}}
\newcommand{\postmean}[1]{\vec{m}_{#1}}
\newcommand{\postcov}[1]{\mat{S}_{#1}}

\newcommand{\plike}{\pcond{\Y}{\f, \likeparam}}
\newcommand{\logpliken}{\log \pcond{\yn}{\fn, \likeparam}}
\newcommand{\qjoint}{q(\f, \u | \llambda)}
\newcommand{\qu}{q(\u | \llambda)}

\newcommand{\qkfstar}{q_k(\fstar | \llambda_k)}  
\newcommand{\qknf}{q_{k(n)}(\fn | \llambda_k )}
\newcommand{\qnf}{q_{(n)}(\fnnodot | \llambda)}

\newcommand{\qfmean}[1]{\vec{b}_{#1}}        % mean of q(f)
\newcommand{\qfcov}[1]{\mat{\Sigma}_{#1}}        % covariance of q(f)
\newcommand{\qnkfr}{q_{k(n)}(\vecf_{(n)\Qr} | \llambda_{k \br} )}

\newcommand{\eq}{Eq.\xspace}
\newcommand{\eqs}{Eqs.\xspace}

\newcommand{\cholK}{\mat{\Phi}} % cholesky lower triangular for cov K
\newcommand{\precK}{\mat{\Lambda}} % precision matrix for cov K

\begin{document}

\title{Scalable Grouped Gaussian Processes via Direct Cholesky Functional Representations}
% The \author macro works with any number of authors. There are two commands
% used to separate the names and addresses of multiple authors: \And and \AND.
%
% Using \And between authors leaves it to LaTeX to determine where to break the
% lines. Using \AND forces a line break at that point. So, if LaTeX puts 3 of 4
% authors names on the first line, and the last on the second line, try using
% \AND instead of \And before the third author name.

\author{%
	Astrid Dahl \\
	The University of New South Wales\\
	\texttt{astridmdahl@gmail.com} \\
	 \And
	Edwin V. Bonilla \\
	CSIRO's Data61 \\%and The University of New South Wales\\
	\texttt{edwin.bonilla@data61.csiro.au} \\
	% \AND
	% Coauthor \\
	% Affiliation \\
	% Address \\
	% \texttt{email} \\
	% \And
	% Coauthor \\
	% Affiliation \\
	% Address \\
	% \texttt{email} \\
	% \And
	% Coauthor \\
	% Affiliation \\
	% Address \\
	% \texttt{email} \\
}

%\input{/Users/astmac/repos/dahl_pubs/macros.tex}

% filepaths
%\newcommand{\bpath}{/Users/astmac/repos/latex_shared}

\maketitle

\begin{abstract}
%We consider multi-task regression models where observations are assumed to be a linear combination of several latent node and weight functions, all drawn from Gaussian process priors that allow nonzero covariance between \emph{grouped} latent functions. Motivated by the problem of developing scalable methods for distributed solar forecasting, we exploit sparse covariance structures where latent functions are assumed to be conditionally independent given a group-pivot latent function. We exploit properties of multivariate Gaussians to construct sparse Cholesky factors directly, rather than obtaining them through iterative routines, and by doing so achieve significantly improved time and memory complexity including prediction complexity that is linear in the number of grouped functions. We test our approach on large multi-task datasets and find that sparse specifications achieve the same or better accuracy than non-sparse counterparts in less time, and improve on benchmark model accuracy.
%
We consider multi-task regression models where observations are assumed to be a linear combination of several latent node and weight functions, all drawn from Gaussian process (GP) priors that allow nonzero covariance between \emph{grouped} latent functions.  We show that when these grouped functions are conditionally independent given a  group-dependent pivot, it is possible to parameterize the prior through sparse Cholesky factors directly, hence avoiding their computation during inference. Furthermore, we establish that  kernels that are multiplicatively separable over input points give rise to such sparse parameterizations naturally without any additional assumptions. Finally, we extend the use of these sparse structures to approximate posteriors within variational inference, further improving scalability on the number of functions.  We test our approach on  multi-task datasets concerning distributed solar forecasting and show that it outperforms several multi-task GP baselines and that our sparse specifications achieve the same or better accuracy than non-sparse counterparts.

\end{abstract}

\section{Introduction}
%\gls{GP} models are a flexible nonparametric Bayesian approach that can be applied to various problems such as regression and classification \citep{Rasmussen2006} and have been extended to numerous multivariate and multi-task problems  \cite{Alvarez2012,Wilson2012,Dahl2018,Bonilla2007,Teh2005,Hensman2014,Gardner2018prodkern} including spatial and spatio-temporal contexts \citep{Cressie2011}. To maintain scalability in multi-task \glspl{GP},  the \gls{GP} priors over latent functions may be constrained to be statistically independent as in  \citep{Wilson2012,Dezfouli2015} or covarying with constrained kernel structures to allow algebraic efficiencies as in \citep{Dahl2018,Bonilla2007}.
\gls{GP} models are a flexible nonparametric Bayesian approach that can be applied to various problems such as regression and classification \citep{Rasmussen2006} and have been extended to numerous multivariate and multi-task problems  including spatial and spatio-temporal contexts \citep{Cressie2011}. Multi-task \gls{GP} methods have been developed along several lines (see e.g.~\cite{Alvarez2012} for a review) including mixing approaches that combine multiple latent univariate \glspl{GP} via linear or nonlinear mixing to predict multiple related tasks \citep{Wilson2012,Dahl2018,Bonilla2007,Teh2005,Hensman2014,Gardner2018prodkern}.  
%
%Recently there has been renewed interest in sparse \gls{GP} models and low-rank approximations to reduce the complexity of posterior estimation  \citep{Cutajar2016,Wilson2015kissgp,Gardner2018gpytorch}, which usually requires expensive algebraic operations such as Cholesky factorization of the Gram matrix. These operations, when computed na\"{i}vely and under no additional assumptions, have a cubic time complexity on the number of observations, hence being an inherent computational limitation to inference in most \gls{GP} methods.
%
To maintain scalability in multi-task mixing models, various constraints have been employed. In particular, latent \glspl{GP} may be constrained to be statistically independent as in  \citep{Wilson2012,Dezfouli2015} or covarying with constrained kernel structures to allow algebraic efficiencies as in \citep{Dahl2018,Bonilla2007}.

In this paper we consider the multi-task setting where subsets of latent functions in Gaussian process regression networks (\gprn; \cite{Wilson2012}) covary within a constrained structure. We build upon the \gls{GGP} approach of \cite{Dahl2018}, where groups of latent functions may covary arbitrarily with a separable kernel structure. Posterior estimation in this \gls{GGP} framework, as originally proposed in \cite{Dahl2018},  is underpinned by variational inference based on inducing variables \cite{Titsias2009} and its stochastic optimization extension \cite{Hensman2013}, hence it should be inherently scalable to a large number of observations.  However, both the time and space complexity deteriorate significantly when considering a large number of tasks, due to the assumed grouping structure. 

Therefore, to address the above limitation, we consider the case where grouped functions are conditionally independent given a specific group-dependent \emph{pivot} function. We exploit this structure in the prior and in the approximate posterior within a variational inference framework  to develop an efficient inference algorithm for the \gls{GGP} model.  Our approach outperforms competitive multi-task baselines and  is significantly more efficient (in time and memory) than its non-sparse counterpart. Our specific contributions are given below.

\textbf{Direct sparse Cholesky functional representation}: We show that when the grouped functions in a \gls{GGP} model are conditionally independent given a  group-dependent pivot, it is possible to parameterize the prior through sparse Cholesky factors directly, hence avoiding their computation during inference. We refer to these factors as \emph{Cholesky functions} as, for a given kernel, they allow us to operate  in the Cholesky factorization space directly as a function of the data and the \gls{GP} hyper-parameters. 

\textbf{Exact construction of sparse Cholesky functionals}: We establish for the first time (to the best of our knowledge)  that  kernels that are multiplicatively separable over input points give rise to such sparse parameterizations naturally without any additional assumptions. This enables \emph{direct construction} of sparse Cholesky factors without resorting to iterative routines that are inherently costly and unstable, potentially increasing the scope for such kernels to be adopted in other machine learning settings and applications.

\textbf{Sparse approximate posteriors}: Finally, we extend the use of sparse structures to the corresponding approximate distributions within a variational inference framework. The required expectations over these  distributions are neatly estimated using a simple but effective `indirect' sampling approach, which further improves scalability on the number of functions. 

\textbf{Experiments}: Our approach is driven by the problem of forecasting solar power output at multiple sites. We apply our \gls{GGP} model and sparse inference framework to this problem using two  real multi-task datasets, showing that it outperforms competitive multi-task benchmarks and  achieve the same or better forecast performance than non-sparse counterparts in significantly less time.

%We also show that the sparse and non-sparse GGP models improve on performance over LCM and GPRN benchmark
%models. While the GPRN performs similarly to GGP models based on root mean squared error (RMSE) and mean
%absolute error (MAE), we show that the GGP models (both sparse and non sparse) achieve significant gains over the
%GPRN on two important metrics, namely the negative log predictive density (reductions range from 8 to 16 percent)
%and forecast variance (reductions range from 22 to 26 per cent).
% Accurate probabilistic forecasting and forecast
%variance are key attributes of solar models as inputs into various physical energy management systems.

\section{Multi-task \gls{GP} regression}
\label{sec:theory_gp}
A Gaussian process (\gls{GP}; \cite{Rasmussen2006}) is formally defined as a distribution over functions such that  $f(\x) \sim \gp (\meanfun(\x), \kernel(\x, \xprime) )$ is a Gaussian process with mean function $\meanfun(\x)$ and covariance function $\kernel(\x, \xprime)$ iff any subset of function values $f(\x_1), f(\x_2), \ldots, f({\x_N})$ follows a Gaussian distribution with mean $\vecS{\mu}$ and covariance $\mat{K}$, which are obtained by evaluating the corresponding mean function and covariance function at the input points $\X = \{\x_1, \ldots, \x_N \}$.

In this paper we consider a form of multi-task \gls{GP} regression where multiple outputs are modeled as a linear combination of node and weight functions, each of which is a \gls{GP}. Data  is of  the form $\dataset = \{\X \in \setR^{\n \times \d},  \Y \in \setR^{\n \times \p} \}$ where each $\xn$ in $\X$ is a $\d$-dimensional vector of input features and each $\yn$ in $\Y$ is a $\p$-dimensional vector of task outputs.  We consider the \acrfull{GGP} model of \cite{Dahl2018} who place a prior over $\q$ latent \gls{GP} functions $\f = \{f_j(\x)\}_{j=1}^\q$ such that arbitrary, non-overlapping subsets (`groups') of latent functions may have nonzero covariance. 

We denote arbitrarily chosen subsets in $\f$ as $\fr \in \setR^{\n \times \Qr}, \br = 1, \ldots, \bR$, 
where $\bR$ is the total number of groups. For each group the number of latent functions within is denoted $\Qr$ (group size) such that $\sum_{\br=1}^{\bR} \Qr = \q$. In the \gls{GP} each group is comprised of latent functions $\fr = \{ f_j \}_{j \in \text{ group } \br} $ and the covariance between two functions is non-zero iff the corresponding processes belong to the same group. This leads to a prior over functions given by
\begin{equation}
\label{eq:priorlnk}
\pcond{\F}{\hyperparam} = \prod_{\br=1}^{\bR} \pcond{\fr}{ \hyperparam_{r}} = \prod_{\br=1}^{\bR} \Normal (\fr ; \vec{0},  \Krxxhh ),
\end{equation}
where $\Krxxhh \in \setR^{\n\Qr \times \n\Qr}$ is the covariance matrix generated by the group kernel function $\krxh$, which evaluates the covariance of functions $f_j$ and $f_{j^\prime}$ at the locations $\x$ and $\xprime$, respectively. $\krxh = 0$ iff  $f_j$ and $f_{j^\prime}$ do not belong to the same group $r$. 

Correlations between outputs are modeled as in the Gaussian process regression network (\gprn) likelihood of \cite{Wilson2012}, where $\fn$ is a $\q$-dimensional vector of latent function values at time $n$ and
\begin{equation}
\plike = \prod_{n=1}^{\n} \pcond{\yn}{\fn, \likeparam}  \nonumber \\
= \prod_{n=1}^{\n} \Normal(\yn; \Wn \gn, \mat{\Sigma}_y) \text{.}
\end{equation}
Here we define $\W$ and $\g$ subsets of $\f$ formed by gathering $\p\Qg$ and $\Qg$ functions in $\f$, respectively, with $\Qg(\p+1) = \q$, $\likeparam = \mat{\Sigma}_y$,  $\fn = \{\Wn, \gn\}$ and  $\mat{\Sigma}_y$ is a diagonal matrix. 
$\p$-dimensional outputs are constructed  at $\xn$ as the product of a $\p \times \Qg$ matrix of weight functions, $\Wn$, and $\Qg$-dimensional vector of node functions $\gn$. 
% in the approach of \citep{Dahl2018}, which builds on the generic inference method allowing `black box' likelihoods of \citep{Dezfouli2015}, 
Partitions of $\f$ with respect to $\W$ and $\g$ need not align with partitions into groups $\fr$. Hence grouping in the prior can be independent of the likelihood definition and, for brevity, inference is presented below simply in terms of $\f$ and $\fr$ rather than $\W$ and $\g$.

To maintain scalability of the approach,  we consider separable kernels of the form $\krxh = \krxx \krhh$, where for each group $\Qr$ vectors $\h \in \dimh$ form a feature matrix $\Hrmat \in \setR^{\Qr \times \dimh}$ that governs covariance across functions $f_j \in \fr$. Group covariance $ \Krxxhh = \Krhh \kron \Krxx$ thus decomposes into $\Krxx \in \setR^{\n \times \n}$ and $\Krhh \in \setR^{\Qr \times \Qr}$. 

In this paper we propose sparse forms of $\Krhh$ arising from a constrained form of cross-function covariance, whereby functions within a group are conditionally independent given a group `pivot' latent function. By exploiting conditional independence constraints that can reasonably fit with spatio-temporal applications such as distributed solar prediction (rationale are discussed in the supplement, \S \ref{sec:expts-setup-details}), it is possible to dramatically reduce the complexity of the \gls{GP}  with respect to group size.  
%This is possible due to properties of multivariate Gaussians that allow direct construction of sparse precision and Cholesky factor matrices as discussed below. We do not propose the same form of sparsity for $\Krxx$ as in the context of spatiotemporal modelling the implied temporal conditional independence constraints in that case become problematic.
%
%Since spatial feature matrix $\Hrmat$ and hence $\Krhh$ are constant across observations $\xn$, we also consider the case where $\Krhh$ is freely parameterized rather than feature-dependent. 

\section{Sparse multi-task model}
\label{sec:sparse_mtm}
In this section we describe the main assumptions on the task-dependent covariance of our sparse \gls{GGP} model that will yield significant computational gains over the original \gls{GGP}. The starting point is that of conditional independence across functions given a group-\emph{pivot} latent function.
\label{sec:mvn}
\subsection{Conditional independence}
\label{sec:sparsity_ci}
When variables are jointly normally distributed and subsets of these variables are conditionally independent, their multivariate normal distribution is known to have certain useful properties. Suppose variables $\fr = f_1, f_2, \ldots, f_{\Qr}$ are jointly normally distributed with covariance $\K$ (subscripts on $\K$ are dropped for ease of exposition), and suppose that, given some variable $f_k$, the remaining variables are conditionally independent. That is,
\begin{equation}
\fr \sim \Normal (\mu, \K) \quad \text{ and } \quad f_{i} \perp f_{j}\cond f_{k}, \qquad \forall \quad i,j \neq k, \quad i \neq j ,
\label{eq_ci_mvn}
\end{equation}
where $f_{i} \perp f_{j} \cond  f_{k}$ denotes independence of $f_{i}$ and $f_{j}$ given $f_{k}$. This joint distribution can be represented as a `star' or `tree' undirected graphical model, where $f_{k}$ can be conceived as a `pivot' variable connecting all other variables (see Figure \ref{fig:spggp_ugm} in the supplement).

Where variables are jointly distributed according to \eqref{eq_ci_mvn}, known, sparse expressions for the Cholesky factor and inverse covariance, that allow direct construction of these matrices, can be obtained analytically  \citep{Sun2005chol,Sun2005prec,Whittaker2009}.  For $i,j \neq k$, $i \neq j$, the covariance element $\K_{i,j} = Cov(f_i, f_j)$ is given by
	\begin{equation}
	\K_{i,j} = \K_{i,k} \K_{k,k}^{-1} \K_{k,j}   \iff 
	f_{i} \perp f_{j} \cond f_{k}   \text{ , }  
	\text{leading to} \quad
	\cholK_{i,j} = 0 \text{ }  
	\text{ and } 
	\precK_{i,j} = 0,
	\label{eq_sparsemats}
	\end{equation}
where $\precK$ and $\cholK$ are the precision matrix and lower triangular Cholesky factor of $\K$.\footnote{This result also holds for the multivariate analogue where, for example, $\fr$ is partitioned into subsets $(\f_{\Qr1}, \f_{\Qr2}, \ldots)$ in which case $\K_{i,j} = Cov(\f_i, \f_j)$ represents a submatrix, and similar for $\cholK_{i,j}$ and $\precK_{i,j}$.}
	
Moreover, nonzero elements of $\precK$ and $\cholK$ have a known form (see \cite{Sun2005chol,Sun2005prec} for a useful summary). Without loss of generality, where $f_{2}, \ldots, f_{\Qr}$ are conditionally independent given $f_{1}$, the precision matrix takes a known winged-diagonal form (see the supplement).
%\begin{align}
%&\precK_{11} = \K_{11}^{-1} + \sum_{i=2}^{\Qr} \K_{11}^{-1} \K_{1i} \K_{ii . 1}^{-1}  \K_{i1}  \K_{11}^{-1}  \nonumber \\
%&\precK_{i1} = - \K_{11}^{-1} \K_{1i} \K_{ii . 1}^{-1}, \quad \text{and} \quad
%\precK_{ii} = \K_{ii . 1}^{-1}, \quad \text{with}  \nonumber \\
%&\K_{ii.1}^{-1} = \K_{ii} - \K_{i1}  \K_{11}^{-1}  \K_{1i}, \quad i=2, \ldots, \Qr
%%\label{eq_precelems}
%\end{align}
%\begin{align}
%\precK_{11} &= \K_{11}^{-1} + \sum_{i=2}^{\Qr} \K_{11}^{-1} \K_{1i} \K_{ii . 1}^{-1}  \K_{i1}  \K_{11}^{-1} , \quad 
%\precK_{1i} = - \K_{11}^{-1} \K_{1i} \K_{ii . 1}^{-1}, \quad
%\precK_{ii} = \K_{ii . 1}^{-1}, \quad  \text{ with } \nonumber \\
%\K_{ii.1}^{-1} &= \K_{ii} - \K_{i1}  \K_{11}^{-1}  \K_{1i}, \quad i=2, \ldots, \Qr .
%%\label{eq_precelems}
%\end{align}
%
Of key importance for our model, the associated Cholesky lower triangular factor $\cholK$ also has a known, sparse form 
%\begin{align}
%\cholK = 
%\begin{bmatrix}
%\cholK_{11} & 0 & 0 & \cdots & 0 \\
%\cholK_{21} & \cholK_{22} & 0 & \cdots & 0 \\
%\cholK_{31}  & 0 & \cholK_{33} & \cdots & 0 \\
%\vdots & \vdots & \vdots & \ddots & \vdots \\
%\cholK_{\Qr1} & 0 & 0 & \cdots & \cholK_{\Qr\Qr}
%\end{bmatrix}
%\label{eq_spggp_chol}
%\end{align}
where
$\cholK_{11} = \textrm{Chol}(\K_{11})$, $\cholK_{i1} = \K_{i1} (\cholK_{11})^{-1}, i=2,\ldots,\Qr$ and 
$\cholK_{ii} = \textrm{Chol}(\K_{ii.1} )$, $ i =2, \dots, \Qr$, and $\K_{ii.1} = \K_{ii} - \K_{i1}  \K_{11}^{-1}  \K_{1i}, \quad i=2, \ldots, \Qr$, and all other elements are zero. $\textrm{Chol}(\cdot)$ denotes the Cholesky factorization of the given argument. 

\subsection{Sparse Gaussian process}
Let $f(\h)$ be drawn from a Gaussian process, $f(\h) \sim \gp(\vec{0}, \kernel (\h, \hprime))$ and assume that
$f(\h_i) \perp f (\h_j) \quad \lvert \quad f(\h_k)$ for some $i,j \neq k, i \neq j$. Thus, the properties of multivariate Gaussians above imply the constrained covariance form
\begin{align}
\kernel(\h_i, \h_j) & \equiv \kernel(\h_i, \h_k) \kernel(\h_k, \h_k)^{-1} \kernel(\h_k, \h_j).
\label{eq_cikern}
\end{align}
Again without loss of generality, setting $k=1$ and $i = 2, \ldots, \Qr$ yields a sparse form of the Cholesky factor where 
\begin{equation}
\cholK_{11}  = \mathrm{Chol}(\kernel_{11}), \quad 
\cholK_{i1} = \kernel_{i1} (\cholK_{11})^{-1}, \quad \text{and}   \quad
\cholK_{ii} = \mathrm{Chol}(\kernel_{ii} - \kernel_{i1} \kernel_{11}^{-1} \kernel_{1i} ).
\label{eq_sparsechol_elems}
\end{equation}
This form has several useful characteristics. Computation involves only univariate operations, meaning that $\mathrm{Chol}(\cdot) = \sqrt{(\cdot)}$, $(\cholK_{11})^{-1} = \frac{1}{\cholK_{11}}$ and $\kernel_{11}^{-1} = \frac{1}{\kernel_{11}}$.
Since there are $2\Qr-1$ nonzero terms, complexity of both computation and storage is $\bigO(\Qr)$ rather than the more general  $\bigO(\Qr^3)$ and $\bigO(\Qr^2)$.  Computation involving univariate operations only also allows direct control over numerical stability of the factorization.

In addition, the sparse form can be decomposed as two sparse matrices with a single nonzero column and nonzero diagonal respectively, \ie
\begin{equation}
\cholK =  %\begin{bmatrix}
[ \cholK_{.1},  \mathbf{0}_1,  \ldots,  \mathbf{0}_{\Qr-1} ] 
%\end{bmatrix}   \\
% &
+ 
\mathrm{diag}(0, \cholK_{22}, \ldots, \cholK_{\Qr\Qr}),
\end{equation}
where $[\cdot]$ is the concatenation of the first column of $\cholK$ ($\cholK_{.1}$)  and $\Qr-1$ zero column vectors with length $\Qr$, and $\mathrm{diag}(\cdot)$ is the diagonal matrix with diagonal elements $(0, \cholK_{22}, \ldots, \cholK_{\Qr\Qr})$.
In practice, this means matrix operations can be evaluated using efficient vector-broadcasting, rather than matrix multiplication, routines.

%\noteAD{
To the best of our knowledge, direct construction of sparse Cholesky factors using explicit expressions as above has not been  employed in Gaussian process models. Rather, where constrained covariance forms such as given at (\ref{eq_ci_mvn}) are used, including in sparse methods discussed at \S \ref{sec:spggp_relatedwork}, it is in the context of prediction on test points that are conditionally independent given some inducing set of variables or latent functions (see e.g.~\cite{Quinonero-Candela2005}). 
%}
%
\subsubsection{Exact implicit sparsity in \gls{GP} priors}
\label{sec:spggp_implicit_spgp}
In some spatio-temporal settings it may be reasonable to explicitly impose the constraint at (\ref{eq_cikern}), which we term `explicit' sparsity.
In other cases, construction of a Gram matrix, associated Cholesky factor and inverse using general routines and using \eqref{eq_sparsemats} -- \eqref{eq_sparsechol_elems} are equivalent. This is due to certain kernels implicitly giving rise to the identity in (\ref{eq_cikern}).
Where kernels can be expressed as products of real-valued functions of $\h_i$ and $\h_j$, \ie assuming 
$\kernel(\h_i, \h_j) = \phi(\h_i) \psi(\h_j)$, and assuming the inverses ($\phi(\h_i)^{-1}$, $\psi(\h_j)^{-1}$) are defined for all $\h_i, \h_j \in \Hmat$, kernels give rise to this identity (see the supplement for details). 
The requirement for symmetry in Mercer kernels  (see \eg \cite{Genton2001}) requires $\kernel(\h_i, \h_j) = \kernel(\h_j, \h_i)$, implying
$\phi(\h_i) \psi(\h_j) = \phi(\h_j) \psi(\h_i)$ for all $\h_i, \h_j \in \Hmat$, %\noteAD{Is it all in $\Hmat$ or all in $\mathcal{H}$?}.
%
%In a univariate \gls{GP} where $\phi(\h)$ and $\psi(\h)$ are identical, this kernel form is a particular case of the Relevance Vector Machine of \cite{Tipping2001}. However, 
however we note that functions ($\phi(\h)$, $\psi(\h)$) need not be identical for (\ref{eq_cikern}) to hold.

Trivially, multiplicative kernels comprised of kernel functions that have the property at (\ref{eq_cikern}) retain this property. Thus Gram matrices that can be expressed as the Hadamard product of matrices constructed via kernel functions of this type also retain the properties at (\ref{eq_sparsemats}). 
Kernels that meet this criterion include constant kernels, polynomial kernels in one dimension, dot product wavelet kernels, and separable stationary kernels (see \citep{Genton2001}). All these kernels decompose multiplicatively into real valued functions of inputs points $\h_i$ and $\h_j$ in a straightforward way. %Other, as yet unidentified, kernels may also give rise to the identity in  (\ref{eq_cikern}).

\subsubsection{Properties of implicitly sparse kernels}
\textbf{Direct decomposition:} 
%Following from \eqref{eq_ci_kern} and \eqref{eq_cikern}, 
When kernels decompose multiplicatively for any points $\h_i$, $\h_j$ and $\h_k$, the Cholesky has the form defined by \eqref{eq_sparsechol_elems}.
The Cholesky can be expressed and directly constructed in this way because the relationship holds for any three points. Therefore \emph{any point can be assigned as the `pivot' point $\kernel_{11}$ and pairwise covariances between any other two points can be expressed in terms of covariances with $\kernel_{11}$}.

\textbf{Degeneracy:} A corollary of this, however, is that where (\ref{eq_cikern}) holds, only one point on the diagonal is nonzero, $\Phi_{11}$, since for all other points, $\kernel_{ii} - \kernel_{i1} \kernel_{11}^{-1} \kernel_{1i} = 0$ (from (\ref{eq_sparsechol_elems}))%. This implies the Cholesky factor consists of a single nonzero column and $\K$ is singular. This in turn implies 
, implying that the \gls{GP} is degenerate.\footnote{The latter follows from the result provided in \citep{QuinRassWilliams2007} that a \gls{GP} is degenerate iff the covariance function has a finite number of nonzero eigenvalues.}
The kernels listed above that decompose multiplicatively are positive semi-definite, as opposed to strictly positive definite, Mercer kernels \citep{Rasmussen2006,Zhang2004}.
One of the criticisms of the degenerate model is that, for distance-based kernels, predictive variance at new points reduces with distance from observed points and in the limit can be zero \citep{QuinRass2005RRGPR}.

\textbf{Avoiding degeneracy in \gls{GGP} models:} Issues around degeneracy are avoided in our framework for two reasons. Firstly, use of a sparse construction is limited to $\Krhh$. In essence, the model induces sparsity over \emph{tasks} rather than \emph{observations}. Under multi-task latent function mixing approaches, including \gls{GGP}, it is generally the case that predictions are not made on new tasks. As such, there is no degeneration over test points since locations across latent functions are fixed for training and test observations. Secondly, as is common practice we add a Kronecker delta kernel (see \citep{Genton2001}) to diagonal elements, however excluding the pivot 
%to diagonal elements $diag( \cholK_{22}, \ldots, \cholK_{\q\q})$, 
which maintains the model as an exact \gls{GP} prior without losing the direct Cholesky construction. This is possible because, for $\Krhh$, the pivot input point does not change. Pivot choice is discussed at \S \ref{sec:solar_application} and in the supplement.
\section{Inference}
\label{sec:inference}
Inference for the sparse \gls{GGP} follows the sparse variational approach of \cite{Titsias2009} and extended by \citep{Dezfouli2015,Dahl2018} where the prior at (\ref{eq:priorlnk}) is augmented with inducing variables $\{ \ur \}_{\br = 1}^{\bR}$, drawn from the same \gls{GP} priors as $\fr$ at $\m$ inducing points $\Zr$ in the same space as $\X$, giving 
\begin{equation}
\label{eq_priorsparse}
\pcond{\u}{ \hyperparam} = \prod_{\br=1}^{\bR} \Normal (\ur ; \vec{0},  \Krzzhh ) \quad  \text{ and } \quad
\pcond{\f}{\u}  = \prod_{\br=1}^{\bR} \Normal (\fr ; \priormeanr, \priorcovr ),  
\end{equation}
where  $\priormeanr = \Ar\ur$,  $\priorcovr = \Krxxhh - \Ar\Krzxh$  and $\Ar = \Krxzh\Krzzhhinv = \I_{Qr} \kron \Krxz\Krzzinv$. 
$\Krzzhh \in \setR^{\m\Qr \times \m\Qr}$ is governed by $\krxh$ evaluated over $\Zr$, $\Hrmat$, similarly yielding the decomposition $\Krzzhh = \Krhh \kron \Krzz$.
 The joint posterior distribution over $\{\f, \u \}$ is approximated by variational inference with
\begin{equation}
\pcond{\f,\u}{\y}   \approx \qjoint \defeq \pcond{\f}{\u}\qu  \quad \text{with} \quad
%\end{equation}
%where the general form of $\qu$ is a mixture of Gaussians,
%\begin{align}
\qu = \sum_{k=1}^{\k} \kprop \prod_{\br=1}^{\br} \qukr ,
\label{postvar}
%\end{align}
\end{equation}
where $\qukr = \Normal(\ur ; \postmean{k\br}, \postcov{k\br})$ and $\varparamkr = \{\postmean{k\br}, \postcov{k\br}, \kprop \}$. $\postmean{k\br}$ and $ \postcov{k\br}$ are, respectively, a freely parameterized mean vector and covariance matrix and $\kprop$ are mixture weights.
Prediction for a new point $\ystar$ given $\xstar$ is computed as the expectation over the variational posterior  for the new point:
\begin{align}
\label{eq:pred-distro}
\pcond{\ystar}{\xstar} &= \sum_{k=1}^{\k} \kprop \int  \pcond{\ystar}{\fstar}\qkfstar \df_{\star} ,
\end{align}
where $\qkfstar$ is defined as for $\qknf$ below (see \S \ref{sec:inf_complexity}). 
The expectation in \eq \eqref{eq:pred-distro} is estimated via \gls{MC} sampling:
$
\expectation{\pcond{\ystar}{\xstar}}{\ystar} \approx \frac{1}{\s} \sum_{s=1}^{\s} \W_{\star}^{s}\vecg_{\star}^{s} ,
$
where $\{ \W_{\star}^{s}\text{,}\vecg_{\star}^{s} \} = \fstar^{s}$ are samples from $\qkfstar$. To estimate the posterior parameters 
we optimize the  \gls{ELBO} defined as 
$ \elbo  \defeq \enterm + \crossterm  + \ellterm$ where $\enterm$, $\crossterm$ and $\ellterm$ are entropy, cross entropy and expected log likelihood terms, respectively. Derivations of the general expressions for the \gls{GP} model and \gls{ELBO} can be found at \citep{Dahl2018,Dezfouli2015}.
In evaluating the efficiency of our approach we consider both a diagonal and Kronecker structure for $\postcov{k\br}$. We define the Kronecker specification as 
$
\postcov{k \br} = \postcovkronb \kron \postcovkronw
$
where $\postcovkronb \in \setR^{\Qr \times \Qr}$ and $\postcovkronw \in \setR^{\m \times \m}$ are both sparse, freely parameterized matrices that heuristically correspond to `between' and `within' covariance components. Sparsity is induced via Cholesky factors of the form at (\ref{eq_sparsechol_elems}).
\subsection{Computational gains and indirect sampling}
\label{sec:inf_complexity}
The corresponding components of the \gls{ELBO} are as follows, 
\begin{align}
& \enterm \geq 
- \sum_{k=1}^{\k} \kprop \log  \sum_{l=1}^{\k} \pi_l   \prod_{\br=1}^{\bR} \Normal (\postmean{k\br}; \postmean{l \br}, \postcov{k \br} + \postcov{l \br} ) \defeq \widehat{\enterm},
\label{eq_ent_main} \\
\crossterm (\llambda) &=  -\frac{1}{2} \sum_{k=1}^{\k} \kprop \sum_{\br=1}^{\bR} \left[ c_r
+ \log \det{\Krzzhh}  
+ \postmean{k\br}'\Krzzhhinv \postmean{k\br} + \trace(\Krzzhhinv \postcov{k\br}) \right] , 
\label{eq_cross_main} \\
\ellterm (\llambda) &= \sum_{n=1}^{\n}\expectation{\qnf}{\logpliken} ,
\label{eq_ell_main}
\end{align}
where $c_r = \m_{\br} \log 2\pi$ and $\widehat{\enterm}$ is used instead of $\enterm$. The main computational gains, following the efficiencies due to the Kronecker factorization, arise from the sparsification of the prior and the approximate posterior. Indeed, we can show that costly algebraic operations on the matrices $\postcov{k\br} $  and $\Krhh$ (obtained from the Kronecker factorization of $\Krzzhh$) such as log determinants and inverses are reduced from $\bigO(Q_r^3)$ to $\bigO(Q_r)$ for the entropy and the cross-entropy terms in \eqs \eqref{eq_ent_main} and \eqref{eq_cross_main}. 

However, for the \gls{ELL} term in \eq \eqref{eq_ell_main} we still need to sample from the marginal posterior $\qnf$ which is done by sampling from component group mixture posteriors \ie $\qnkfr = \Normal ( \qfmean{k\br (n)}, \qfcov{k\br (n)} )$ with mean and covariance expressions given by 
$\qfmean{k\br (n)} = \Arn \postmean{k\br}$ and 
$\qfcov{k\br (n)}  = \priorcovrn + \Arn\postcov{k\br}\Arn^{\prime}$, where
$\priorcovr$ and $\Ar$ are defined as in \eqref{eq_priorsparse} (detailed expressions are provided in the supplement, \S \ref{sec:complexity}). 
Na\"{i}vely, this is $\bigO(Q_r^3)$. We address this issue by `indirect sampling', \ie drawing independent samples from two distributions specified as $\Normal(\vec{0}, \priorcovrn)$ and $\Normal(\qfmean{k\br (n)}, \Arn\postcov{k\br}\Arn^{\prime})$, and summing these to return samples from $\Normal ( \qfmean{k\br (n)}, \qfcov{k\br (n)} )$, hence reducing the time complexity to $\bigO(Q_r)$. Similarly, these savings for the \gls{ELL} apply to predictions as obtained from \eq \eqref{eq:pred-distro}. Finally, there are significant gains  from the proposed sparsification and indirect sampling method  in terms of memory complexity, going from $\bigO(Q_r^2)$ to $\bigO(Q_r)$. Full  complexity analysis is given in the supplement, \S \ref{sec:complexity}.

\section{Application to solar \gls{PV} forecasting}
\label{sec:solar_application}
We test the sparse \gls{GGP} model on solar forecasting applications, where tasks are solar sites and the modelling goal is to jointly predict power output at all sites at each test time point $n$. 
The sparse \gls{GGP} model for solar follows the approach in \citep{Dahl2018}  where, for $\p$ tasks, there are $\p$ latent node functions, meaning $\Qg(\p+1) = \q$. Latent weight functions are grouped according to rows of $\Wn$ while latent node functions are assumed to be independent. Thus, predicted output at site $i$, $\vecy_{(n)i}$, is a linear combination of (site-associated) node functions. Spatial features, $\h_j = (latitude_{j}, longitude_{j}), j=1,\ldots,\Qr$, populate the cross function feature matrix $\Hrmat$ for every group. We test three forms of $\Krhh$ Cholesky factors: an implicitly sparse form via multiplicatively separable kernels; an explicitly sparse form where the equality in (\ref{eq_cikern}) is imposed rather than naturally arising; and a sparse, freely parameterized factor. Group pivot functions are set as the diagonal elements of $\Wn$ (see the supplement, \S \ref{sec:expts-setup-details}).

%Setting $\p = \Qg$, meaning $\Wn$ is a square matrix, and grouping latent functions by rows allows spatial features across sites to be used in $\Krhh$ to spatially smooth weights applied by each site across node functions. Because row-groups are independent from each other, each task has uniquely parameterized, smoothed latent weights. 
%
%Kernels and features for $\krxx$ and $\krhh$ follow  \citep{Dahl2018}.
% 
\subsection{Experiments}
\label{sec:spggp_experiments}
Data are five minute average power output between 7am and 7pm. We test our model using two datasets (with $\p=25$ and $\p=50$).\footnote{Datasets and implementations are available on GitHub. See the supplement for details.} The first dataset consists of 25 proximate sites over 45 days in Spring with 105,000 (57,000) observations for training (testing). %Sites are located within an approximately 20 by 20 kilometre area.
The second dataset consists of 50 proximate sites 
%within an approximately 30 by 30 kilometre area 
in a different location and season (Autumn) with 210,000 (114,000) observations in total for training (testing). We forecast power output 15 minutes ahead at each site over the test period.

For $\p = 25$ we evaluate performance relative to the general \gls{GGP}  and  three classes of non-\gls{GGP} benchmarks: the linear coregional model (\lcm), \gprn and multi-task model with task-specific features (\mtg) of \citep{Bonilla2008}. In addition, we test a general \gls{GGP} model with full, freely parameterized cross site covariance. 
For $\p=50$, we evaluate sparse models using Kronecker posteriors against non \gls{GGP} benchmarks. However, other \gls{GGP} models could not be estimated under the same computational constraints. %All models a single Gaussian approximate posterior, \ie $\k=1$.
We consider the \gls{RMSE}, \gls{MAE}, \gls{NLPD}, \gls{MRANK} and \gls{FVAR} to evaluate the performance of all methods. 
Full details of the experimental set-up are given in the supplement, \S \ref{sec:expts-setup-details}. 
\subsubsection{Results}
Results for all models are reported at Table \ref{tab:spggp_results}with further results available in the supplement (\S \ref{sec:additional-expts}). 
For $\p=25$ forecast accuracy of sparse and general \gls{GGP} models is comparable. Results for sparse and non sparse \gls{GGP} models differ by less than a percentage point on \rmse and \mae, while \nlpd results are slightly more variable. 
% Differences between diagonal and Kronecker posteriors are also fine although  Most measures for \gls{GGP} and sparse \gls{GGP} models are slightly improved under Kronecker posteriors. 
% Table generated by Excel2LaTeX from sheet 'latex_tabs'
\begin{table}[t]
	\small
  \centering
  \caption{Forecast accuracy and variance of \gls{GGP}, sparse \gls{GGP} and benchmark models. Results shown for Kronecker posteriors with $\k=1$ for Adelaide ($\p=25$) and Sydney ($\p=50$) datasets. Results are reported for best performing \gprn and \lcm benchmarks (based on \rmse) and for \lcm where $\Qg=\p$. %\mrank is average model rank over accuracy measures (\rmse, \mae and \nlpd).
  	All metrics are losses, \ie the lower the better.
  }
    \begin{tabular}{lrrrrr}
    	\toprule
    	P = 25 &       &       &       &       &  \\
    	\textit{Kronecker} & \multicolumn{1}{l}{\rmse} & \multicolumn{1}{l}{\mae} & \multicolumn{1}{l}{\nlpd} & \multicolumn{1}{l}{\mrank} & \multicolumn{1}{l}{\fvar} \\
    	\midrule
%    	\textit{Diagonal} &       &       &       &       &  \\
%    	\gls{GGP}   & 0.343 & 0.216 & 0.399 & 7.3   & 0.121 \\
%    	\gls{GGP} (free) & 0.345 & 0.215 & 0.378 & 7.0   & 0.116 \\
%    	Sparse \gls{GGP} (implicit) & 0.344 & 0.217 & 0.423 & 10.7  & 0.124 \\
%    	Sparse \gls{GGP} (explicit) & 0.345 & 0.215 & 0.382 & 7.7   & 0.115 \\
%    	Sparse \gls{GGP} (free) & 0.343 & 0.215 & \textbf{0.371} & 4.7   & 0.115 \\
%    	\lcm ($\Qg=\p$) & 0.371 & 0.235 & 0.553 & 15.0  & 0.181 \\
%    	\lcm ($\Qg=5$) & 0.366 & 0.239 & 0.481 & 14.3  & 0.151 \\
%    	\gprn ($\Qg=2$) & 0.344 & 0.217 & 0.451 & 10.7  & 0.149 \\
%    	\mtg   & 0.381 & 0.241 & 0.501 & 16.7  & 0.171 \\
%    	&       &       &       &       &  \\
    	 &       &       &       &       &  \\
    	\gls{GGP}   & 0.343 & 0.213 & 0.382 & 4.3   & 0.117 \\
    	\gls{GGP} (free) & 0.344 & \textbf{0.213} & 0.384 & 5.3   & 0.112 \\
    	Sparse \gls{GGP} (implicit) & 0.342 & 0.213 & 0.414 & 4.7   & 0.120 \\
    	Sparse \gls{GGP} (explicit) & \textbf{0.341} & 0.214 & 0.374 & \textbf{2.7} & 0.114 \\
    	Sparse \gls{GGP} (free) & 0.344 & 0.216 & 0.378 & 7.0   & \textbf{0.111} \\
    	\lcm ($\Qg=\p$) & 0.375 & 0.236 & 0.583 & 16.0  & 0.180 \\
    	\lcm ($\Qg=4$) & 0.367 & 0.240 & 0.475 & 14.7  & 0.147 \\
    	\gprn ($\Qg=2$) & 0.342 & 0.214 & 0.446 & 5.7   & 0.150 \\
    	\mtg   & 0.381 & 0.237 & 0.502 & 16.3  & 0.170 \\
    	&       &       &       &       &  \\
    	\midrule
    	P = 50 &       &       &       &       &  \\
    	\textit{Kronecker} & \multicolumn{1}{l}{\rmse} & \multicolumn{1}{l}{\mae} & \multicolumn{1}{l}{\nlpd} & \multicolumn{1}{l}{\mrank} & \multicolumn{1}{l}{\fvar} \\
    	\midrule
    	Sparse \gls{GGP} (implicit) & \textbf{0.421} & \textbf{0.254} & \textbf{0.622} & \textbf{1.0} & 0.159 \\
    	Sparse \gls{GGP} (explicit) & 0.421 & 0.257 & 0.626 & 2.3   & 0.141 \\
    	Sparse \gls{GGP} (free) & 0.423 & 0.258 & 0.625 & 2.7   & \textbf{0.139} \\
    	\lcm ($\Qg=\p$) & 0.451 & 0.283 & 0.807 & 5.3   & 0.211 \\
    	\gprn ($\Qg=2$) & 0.428 & 0.263 & 0.664 & 4.0   & 0.185 \\
    	\mtg   & 0.483 & 0.297 & 0.741 & 5.7   & 0.211 \\
    	\bottomrule
    \end{tabular}%
  \label{tab:spggp_results}%
\end{table}%
For $\p=25$, accuracy is comparable for \gls{GGP} and \gprn on \mae and \rmse, however \gls{GGP} based models perform significantly better than benchmarks on \nlpd. The \lcm and \mtg perform relatively poorly on all measures.
Mean forecast variance was also found to be significantly lower (28 percent on average) under \gls{GGP} models relative to benchmark models.
For $\p=50$, sparse \gls{GGP} models outperform benchmark models on all accuracy and variance measures.

\begin{figure}
	\centering
	\begin{tabular}{l l l l l}
		\hspace{7.6mm}Diagonal & \hspace{0.8mm}Kronecker & \hspace{7.6mm}Diagonal & \hspace{0.8mm}Kronecker & \\
	%	\multicolumn{2}{c}{	\includegraphics[width=0.33\linewidth]{timedres/batchtime.png} } &
		\multicolumn{2}{c}{	\includegraphics[width=0.33\linewidth]{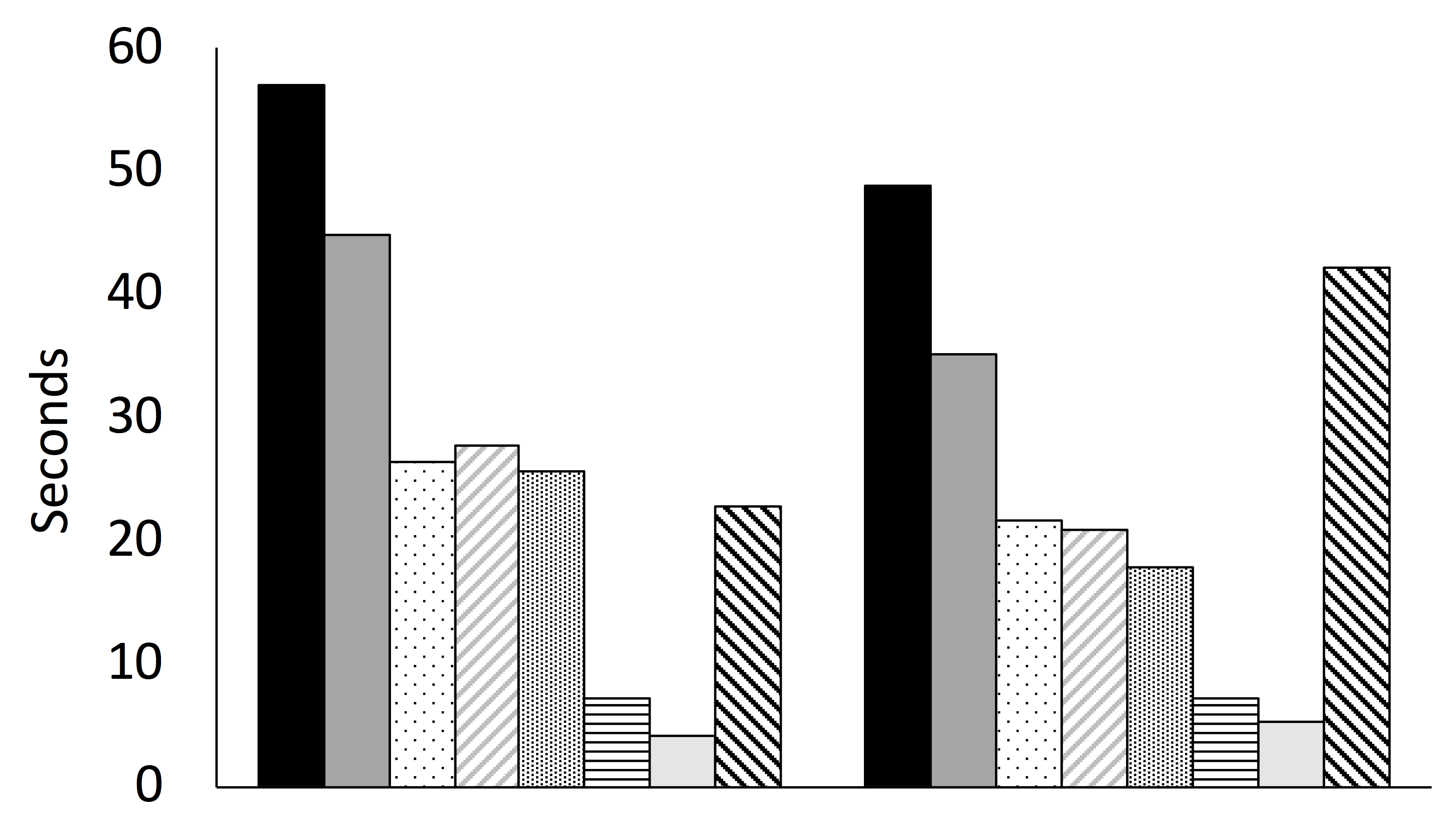} } &
		\multicolumn{2}{c}{	\includegraphics[width=0.33\linewidth]{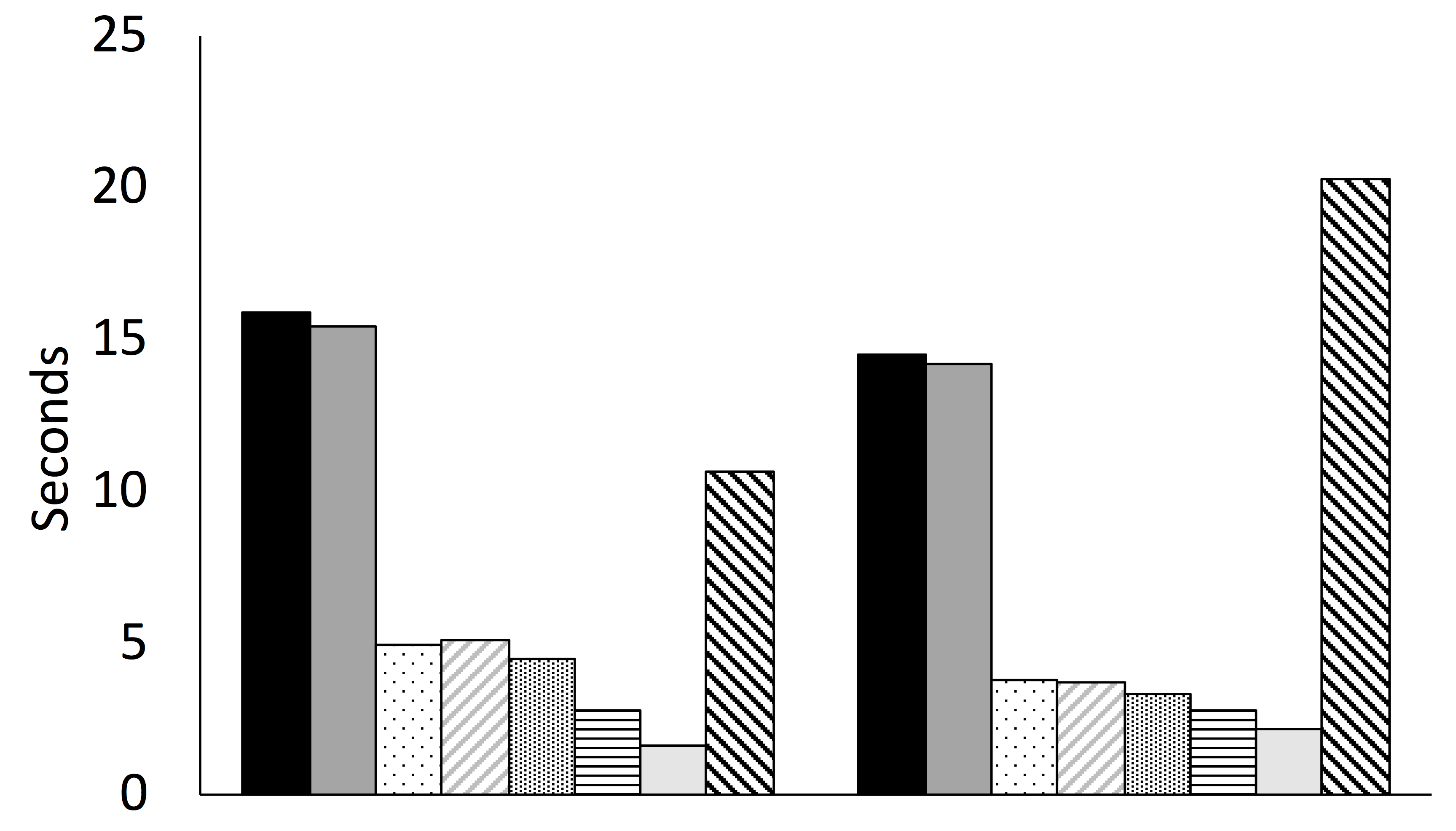} } &
		\includegraphics[width=0.2\linewidth]{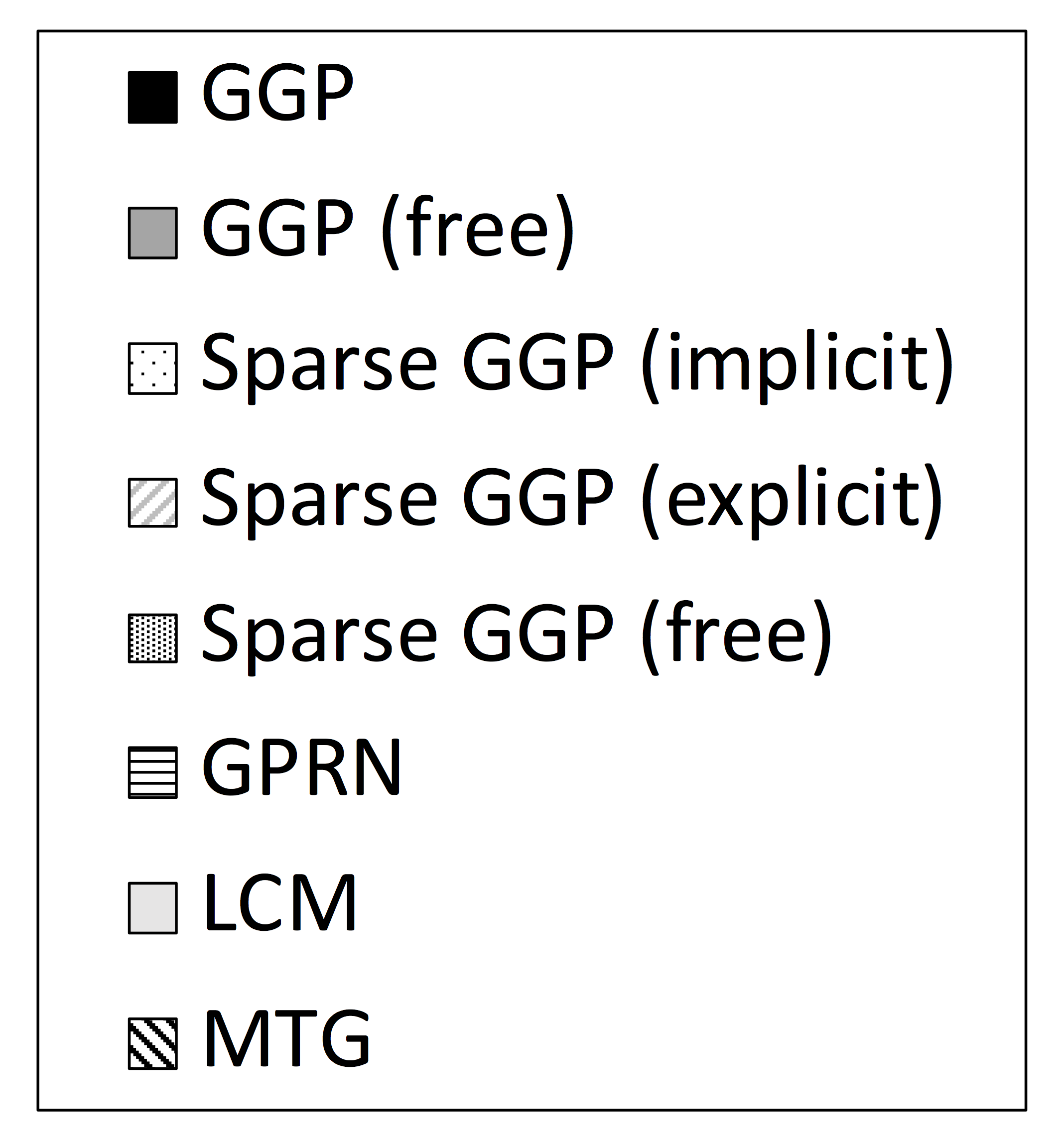} 
%		
%		\&
%		 \\
%		\multicolumn{2}{c}{(c)}	& 	\multicolumn{2}{c}{}	 \\
	\end{tabular}
	\caption{Computation times for \gls{GGP} versus sparse \gls{GGP} models ($\p=25$) for diagonal and Kronecker posteriors. 
		%Panel (a) shows average time for a single minibatch step. 
		\emph{Left:} Average time to compute the full evidence lower bound. \emph{Right:} Average time to compute predictions over the full test set. %Results for diagonal (Kronecker) posteriors are shown in left (right) group for each panel.
	}
	\label{fig:spggp_runtime}
\end{figure}

Computation times are shown at Figure \ref{fig:spggp_runtime} for sparse versus general \gls{GGP} models and benchmark models.
Results confirm that sparse \gls{GGP} models have substantially lower time costs than general counterparts; step time and \gls{ELBO} computation time are decreased by 49 to 57 percent and 43 to 57 percent respectively, with Kronecker posteriors showing greater reductions than diagonal posteriors (55 versus 50 percent, respectively, on average). 
The most substantial improvements, however, are for prediction, where sparse models are three to four times faster over the full test set. Further, mean sparse model prediction time scales close to linearly between $\p=25$ (at 3.6 seconds) and $\p=50$ (at 8.5 seconds). Computation time for step and \gls{ELBO} for $\p=50$ (not shown) scales at a higher-than-linear rate overall (on average three times the cost of comparable models for $\p=25$), which we mainly attribute to the grouping structure selected.

We also find that, for the same prior, average time cost is always lower under the Kronecker posterior, consistent with their lower complexity as discussed in the supplement, \ref{sec:complexity}. Further, freely parameterized \gls{GGP} models have lower time costs than `standard' counterparts reflecting the lower complexity of operations on $\Krhh$ elements ($\bigO(1)$ versus $\bigO(\dimh)$ for explicitly defined kernels).
%
% panel of heatmap figures for sparse ggp
%
%\begin{figure}
%	\centering
%	\begin{tabular}{llll}
%		\hspace{7.6mm}Diagonal & \hspace{0.8mm}Kronecker & \hspace{7.6mm}Diagonal & \hspace{0.8mm}Kronecker \\
%		\multicolumn{2}{c}{	\includegraphics[width=0.33\linewidth]{timedres/batchtime.png} } &
%		\multicolumn{2}{c}{	\includegraphics[width=0.33\linewidth]{timedres/nelbotime.png} } \\
%		%\includegraphics[width=0.4\linewidth]{figures/spggp_runtime/batchtime.png}  &  
%		%\includegraphics[width=0.4\linewidth]{figures/spggp_runtime/nelbotime.png} \\
%			\multicolumn{2}{c}{(a)}	& 	\multicolumn{2}{c}{(b)}	 \\
%		
%		\multicolumn{2}{c}{	\includegraphics[width=0.33\linewidth]{timedres/predtime.png} } &
%		\multicolumn{2}{c}{	\includegraphics[width=0.2\linewidth]{timedres/legend.png} } \\
%		%\includegraphics[width=0.4\linewidth]{figures/spggp_runtime/batchtime.png}  &  
%		%\includegraphics[width=0.4\linewidth]{figures/spggp_runtime/nelbotime.png} \\
%		\multicolumn{2}{c}{(c)}	& 	\multicolumn{2}{c}{}	 \\
%	\end{tabular}
%	%
%	\caption{Computation times for \gls{GGP} versus sparse \gls{GGP} models ($\p=25$). Panel (a) shows average time for a single minibatch step. Panel (b) shows average time to compute the full evidence lower bound. Panel (c) shows average time to compute predictions over the full test set. Results for diagonal (Kronecker) posteriors are shown in left (right) group for each panel.}
%	\label{fig:spggp_runtime}
%\end{figure}

Time costs for \gprn and \lcm benchmarks are lower again than sparse \gls{GGP} based models, in the order of 1.3 to 6.3 times faster per operation. The \mtg, however, has very low step time (up to 27 times faster than sparse \gls{GGP} models) and comparable time cost for \gls{ELBO} in the diagonal case but substantially poorer results for \gls{ELBO} and prediction computation times for remaining tests.

\section{Related work}
\label{sec:spggp_relatedwork}
The problem of forecasting local solar output in the short term is of significant interest for the purpose of distributed grid control and household energy management \citep{Voyant2017a,Widen2015}. A number of studies confirm that exploiting spatial dependencies between proximate sites can yield significant gains in forecast accuracy and stability \citep{Dahl2018,Yang2018}. More importantly, inherent to this application is the need for modeling uncertainty in a flexible and principled way (see \eg \cite{Antonanzas2016}), hence our motivation to use \glspl{GP}.

%Numerous earlier methods sought to exploit sparsity in precision or Cholesky matrices by inducing conditional independence constraints. In general these methods approximate the full covariance matrix for a positive definite kernel with 
%$
%\Krxx \approx \K_{\x, \xtilde} \K_{\xtilde, \xtilde}^{-1} \K_{\xtilde, \x}
%$
%where $\xtilde$ may be some subset of inputs or inducing points, potentially with some diagonal correction (also termed `preconditioning') to ensure nondegeneracy, as in early sparse methods such as FITC or PITC \citep{Snelson2006,Quinonero-Candela2005}. Later studies build on the PITC approach to extend conditional independence concepts to convolutional multi-task models and the variational inference setting \citep{Alvarez2009,Alvarez2011}.
%In fact, in the univariate case the kernel formulation for explicit sparsity in \S \ref{sec:sparse_mod} corresponds to the effective FITC prior \citep{QuinRassWilliams2007}.
%Since the development of sparse variational \gls{GP} methods as in \cite{Titsias2009,Hensman2015a} a number of recent works have sought to reduce complexity of \gls{GP} models by inducing sparsity in covariance or precision matrices. These approaches include assuming conditional independence between functions at points or subsets of points, imposing structure on input points and low rank approximations, and more recently combinations of these ideas.
The literature on scalable inference for \gls{GP} models is vast and the reader is referred to \cite{liu2018gaussian} for a comprehensive review of this area. Here we focus on closely related approaches to ours. Early methods adopt low-rank decompositions or assume some kind of conditional independence \citep{Snelson2006,Quinonero-Candela2005}, with  later studies extending these conditional independence ideas to  convolutional multi-task models \citep{Alvarez2009,Alvarez2011}. Nowadays it is widely recognized that the inducing variable framework of \cite{Titsias2009}, made scalable by \cite{Hensman2013}, is one of the de facto standards for scalable \glspl{GP}. However, other approaches such as those based on random feature expansions of the covariance function remain relevant \cite{hensman2017variational,cutajar2017random}. Another idea that has been used in sparse methods is that of structured inputs, \eg regular one-dimensional gridded inputs, which induce Toeplitz matrices with sparse inverses \cite{Wilson2015kissgp}.
  %The approach in \citep{Wilson2015kissgp} exploits this in combination with low rank approximations to develop fast, conjugate gradient based methods.  

Other methods make use of conditional independence relationships arising naturally in Gaussian Markov random fields, particularly the \gls{INLA} of \cite{Rue2009}. These approaches assume a more general conditional independence structure arising from the Markov blanket property. % (as do \citep{Alvarez2009,Alvarez2011}). As in the approach proposed here, this property induces sparsity in the precision matrix and Cholesky factor. 
Closely related to methods inducing conditional independence, several recent techniques consider low-rank approximations in inducing point frameworks \cite{Gardner2018gpytorch,Cutajar2016precond}. In fact, \cite{Gardner2018gpytorch} use a low-rank pivoted Cholesky approximation with pre-conditioning % in combination with iterative methods to estimate key terms appearing in the objective function 
for various \gls{GP} models including sparse methods. Other approaches using low-rank approximations based on spectral methods have also been proposed as in \cite{Solin2014,Evans2018}.
%
%with $\m$ potentially larger than $\n$. 

While sparse variational and other recent low-rank \gls{GP} methods have provided substantial gains in scalability in recent years, as general approaches they do not necessarily exploit efficiencies where sparse, cross-task covariances can be specified \apriori as may be possible in well understood spatio-temporal problems. Finally, given the increasing  capabilities of current GPU-based computing architectures, it is not surprising to see very recent developments to solve  \gls{GP} regression exactly in very large datasets  \cite{wang2019exact}. 
\section{Discussion}
\label{sec:discussion}
We have shown that by exploiting known properties of multivariate Gaussians and conditional independence across latent functions, it is possible to significantly improve the scalability of multi task mixing models with respect to the number of functions.
We have applied this approach to the problem of solar forecasting at multiple distributed sites using two large multi task datasets and demonstrated that both spatially driven and freely parameterized sparse cross function covariance structures can be exploited to achieve faster learning and prediction without sacrificing forecast performance. In particular the cost of prediction is dramatically reduced and shown to be linear in the number of grouped latent functions. 

%While our experimental implementation incorporates most efficiencies immediately implied by the sparse model, it does not incorporate possible efficiencies for two key trace terms. Hence, we do not claim full linearity in time and memory complexity on the number of tasks. However, we emphasize that in principle these are already available in the proposed model and inference approach, and become attractive as the number of latent functions grows.

%We focus our efforts on efficient structures for cross site covariance, however we note that this is not incompatable with further efficiencies that might be considered to reduce complexity on the number of inducing points, including sparse covariance, structured inputs and low rank approximations. The approach can be seen as an avenue to extend sparse models to multi task mixing frameworks.

\pagebreak
\appendix

\section{An example of a sparse \gls{GGP} prior}
\begin{figure}[h!]
	\centering
	\includegraphics[width=0.5\linewidth]{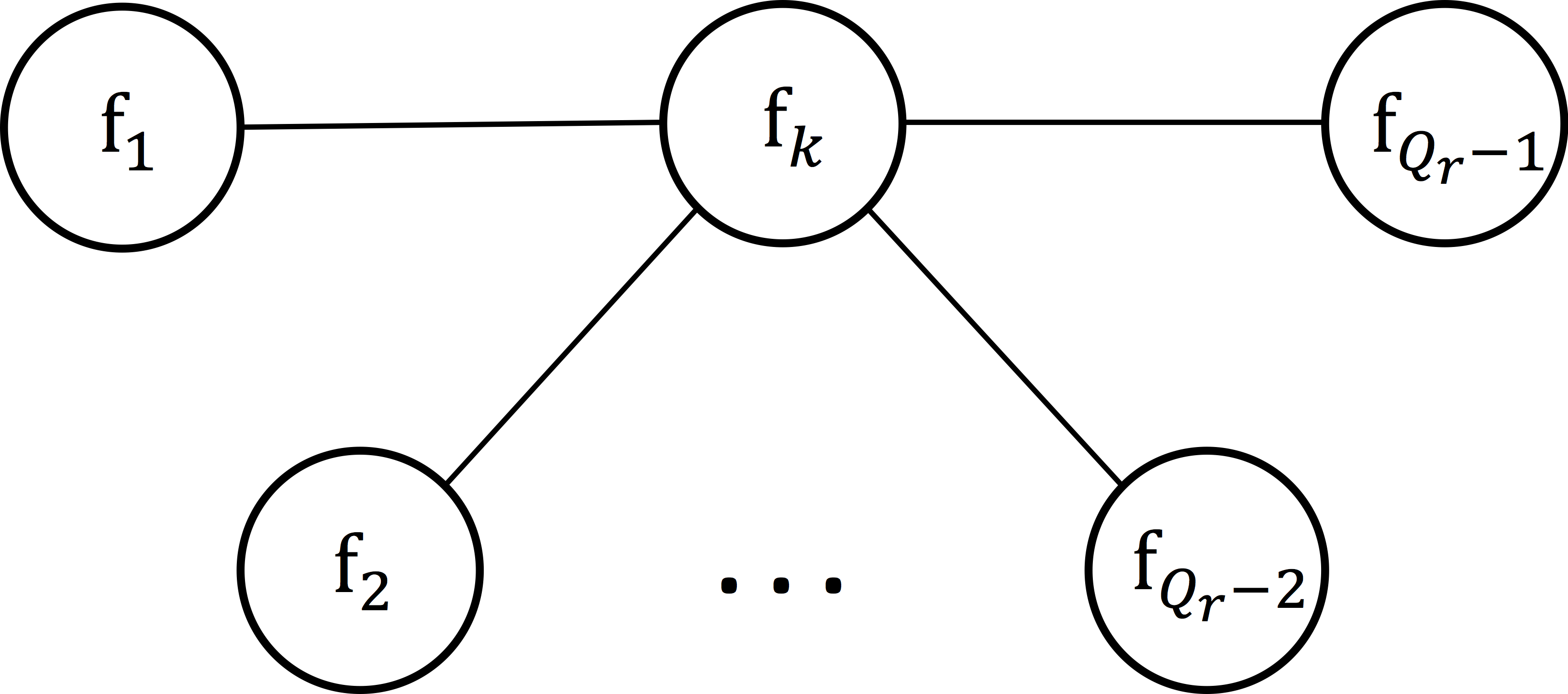}
	\caption{Example of Undirected Graphical Model in Star Formation for $\fr = (f_{1}, f_{2}, \ldots, f_{\Qr})'$.}
	\label{fig:spggp_ugm}
\end{figure}
%
%\onecolumn
\section{Code and Datasets}
We have made our implementation of the sparse  and non sparse \gls{GGP} models available at the Github repository:  \url{https://github.com/spggp/sparse-ggp}.

The repository also contains datasets used in reported experiments and example scripts for execution.
\section{Expressions for multivariate Gaussian precision matrix}
The precision matrix from \citep{Sun2005prec,Whittaker2009}, has a sparse, winged diagonal form
\begin{align*}
\precK = 
\begin{bmatrix}
\precK_{11} & \precK_{21}' & \precK_{31}' &\cdots & \precK_{\Qr1}' \\
\precK_{21} & \precK_{22} & 0 & \cdots & 0 \\
\precK_{31} & 0 & \precK_{33} & \cdots & 0 \\
\vdots & \vdots & \vdots & \ddots & \vdots \\
\precK_{\Qr1} & 0 & 0 &  \cdots & \precK_{\Qr\Qr}
\end{bmatrix} 
\end{align*}
where 
\begin{align*}
\precK_{11} &= \K_{11}^{-1} + \sum_{i=2}^{\Qr} \K_{11}^{-1} \K_{1i} \K_{ii . 1}^{-1}  \K_{i1}  \K_{11}^{-1}  \\
\precK_{1i} &= - \K_{11}^{-1} \K_{1i} \K_{ii . 1}^{-1} \\
\precK_{ii} &= \K_{ii . 1}^{-1}, \quad \text{where} \\
\K_{ii.1} &= \K_{ii} - \K_{i1}  \K_{11}^{-1}  \K_{1i}, \quad i=2, \ldots, \Qr
\end{align*}
The sparse Cholesky factor of $\K$ has the form
\begin{align*}
\cholK = 
\begin{bmatrix}
\cholK_{11} & 0 & 0 & \cdots & 0 \\
\cholK_{21} & \cholK_{22} & 0 & \cdots & 0 \\
\cholK_{31}  & 0 & \cholK_{33} & \cdots & 0 \\
\vdots & \vdots & \vdots & \ddots & \vdots \\
\cholK_{\Qr1} & 0 & 0 & \cdots & \cholK_{\Qr\Qr}
\end{bmatrix}
%\label{eq_spggp_chol}
\end{align*}
where
\begin{align*}
\cholK_{11} & = \mathrm{Chol}(\kernel_{11}), \quad % \nonumber \\
\cholK_{i1} = \kernel_{i1} (\cholK_{11})^{-1}, \quad \text{and}   \nonumber \\
\cholK_{ii} &= \mathrm{Chol}(\kernel_{ii} - \kernel_{i1} \kernel_{11}^{-1} \kernel_{1i} ).
%\label{eq_sparsechol_elems}
\end{align*}

\section{Proof of implicit sparsity in exact \gls{GP} priors}
Construction of a Gram matrix, associated Cholesky factor and inverse using general routines and direct sparse constructions are equivalent where kernels implicitly give rise to the identity where
\begin{align}
\kernel(\h_i, \h_j) &= \kernel(\h_i, \h_k) \kernel(\h_k, \h_k)^{-1} \kernel(\h_k, \h_j) \nonumber \\
 &\iff 
f(\h_i) \perp f(\h_j) \quad \lvert \quad f(\h_k)
\end{align}
for any points $\h_i$, $\h_j$ and $\h_k$.

Where kernels can be expressed as products of real-valued functions of $\h_i$ and $\h_j$, \ie assuming 
$\kernel(\h_i, \h_j) = \phi(\h_i) \psi(\h_j)$, and assuming the inverses ($\phi(\h_i)^{-1}$, $\psi(\h_j)^{-1}$) are defined for all $\h_i, \h_j \in \Hmat$, kernels give rise to this identity. To see this, consider a positive semi-definite kernel such that
\begin{align}
\kernel(\h_i, \h_j) &= \phi(\h_i) \psi(\h_j)  \nonumber \\
\implies \kernel(\h_k, \h_k)^{-1} &= \phi(\h_k)^{-1} \psi(\h_k)^{-1} \nonumber \\
\implies \kernel(\h_i, \h_j) &= \phi(\h_i) \psi(\h_k) \phi(\h_k)^{-1} \psi(\h_k)^{-1} \nonumber \\
                                       & \times \phi(\h_k)  \psi(\h_j) \nonumber \\
&= \kernel(\h_i, \h_k) \kernel(\h_k, \h_k)^{-1} \kernel(\h_k, \h_j)
\end{align}
Kernels of this form are valid, positive semi-definite kernels so long as the properties of symmetry and positive semi-definiteness are satisfied. Symmetry implies $\kernel(\h_i, \h_j) = \kernel(\h_j, \h_i)$, implying
$\phi(\h_i) \psi(\h_j) = \phi(\h_j) \psi(\h_i)$ for all $\h_i, \h_j \in \Hmat$. Positive semi-definiteness implies for any examples $\h_i, \h_j $ and any set of real numbers $\lambda_1, \ldots, \lambda_l$, 
$
\sum_{i=1}^{l} \sum_{j=1}^{l} \lambda_i, \lambda_j \kernel(\h_i, \h_j) \geq 0
$
 (see \eg \cite{Genton2001}). %\noteAD{Is it all in $\Hmat$ or all in $\mathcal{H}$?}.

%\section{Exactness of sparse \gls{GP} models}
%The wavelet kernel over $\h_j = (latitude_{j}, longitude_{j})$ is defined as
%%
%\begin{align}
%\krhh &= \prod_{d \in \h} \left[
%\psi \left( \frac{\h_{jd} - c_d^{\br}}{l_d^{\br}} \right) 
%\psi \left( \frac{\h_{j'd} - c_d^{\br}}{l_d^{\br}} \right) 
%\right] 
%\end{align}
%%
%where $\psi(z)$ is the mother wavelet function. We use the Ricker (Mexican hat) mother wavelet, defined as $(1-z^2) \exp (-0.5 z^2) $. The full kernel is given by 
%$\ksub{\br (wav.)} + \ksub{\br (diag)}$ and it can be easily ascertained that the (full rank) sparse Cholesky can be constructed as in \eqref{eq_spggp_chol} with $\cholK_{ii} = \sqrt{\ksub{\br (diag)}}, i \neq 1$. The diagonal constant is assumed to be shared across all groups.  

We note that following diagonal correction the implicitly sparse model is no longer invariant to pivot choice, however is still an exact \gls{GP} when used as in our model (\ie to characterize cross function covariance in the mixing model). For a collection of random variables to meet the definition of a \gls{GP}, it is only required that any finite subset of the collection is jointly Gaussian distributed, not precluding finite index sets \citep{Rasmussen2006}. 
This observation also extends to the explicitly sparse case and undefined, freely parameterized case.
\section{Computational Complexity}
\label{sec:complexity}
We consider complexity of key terms required for inference and prediction for sparse versus non-sparse models in the following sections, specifically complexity per group of latent functions.
In the discussion that follows, it is assumed that $d$-dimensioned matrix inverses can be evaluated with complexity $\bigO(d^3)$ in the general case and $\bigO(d)$ in sparse or diagonal cases due to nonzero elements of sparse matrices growing linearly with $d$ (see \S \ref{sec:sparsity_ci}).
%It is also assumed that $\Qr \leq \m$. 
%
\paragraph{Entropy}
The entropy component of the \gls{ELBO} is approximated by
\begin{align}
&\enterm \geq 
- \sum_{k=1}^{\k} \kprop \log  \sum_{l=1}^{\k} \pi_l \Normal(\postmean{k}; \postmean{l}, \postcov{k} + \postcov{l}) 
\quad \text{with} \nonumber \\
&\Normal(\postmean{k}; \postmean{l}, \postcov{k} + \postcov{l})
= \prod_{\br=1}^{\bR} \Normal (\postmean{k\br}; \postmean{l \br}, \postcov{k \br} + \postcov{l \br} ).
\label{eq_ent}
\end{align}
The normal terms in (\ref{eq_ent}) differ in complexity over posterior forms and whether $l=k$. In the case where $\k >1$ and with a Kronecker posterior, both the general and sparse \gls{GGP} have poor scalability since evaluation requires the log determinant and inverse of $\postcov{k \br} + \postcov{l \br}$ with complexity $\bigO((\m\Qr)^3)$. Hence, we consider $\k=1$ or diagonal posteriors for $\k \geq1$. In the diagonal case time and space complexity is $\bigO(\m\Qr)$ for sparse and non sparse models ($\bigO(\k\m\Qr)$ for $\k>1$).

For the non-diagonal case, 
%$l=k \implies \postmean{k} = \postmean{l}$, hence 
the entropy component for each group $\br$ reduces to 
$
-\frac{1}{2} (\text{log} \det{2 \postcov{k\br}} ) - C
$
where $C$ is constant with respect to model parameters. Since the log determinant decomposes as 
$
\text{log} \det{2 \postcov{k\br}} = \Qr \m \ln 2 + \m \ln \det{\postcovkronb} + \Qr \ln \det{\postcovkronw}
$ 
%the only difference in estimation of entropy under the \gls{GGP} and sparse \gls{GGP} specifications is in the storage of $\postcovkronb$ and $\postcovkronw$ since in both cases 
evaluation depends only on the diagonal with complexity $\bigO(\Qr + \m)$. The cost of storage is $\bigO(\Qr + \m)$ versus $\bigO(\Qr^2 + \m^2)$ for sparse and general cases.

\paragraph{Cross Entropy}
$\crossterm$ has several components that differ across sparse and general models, since
\begin{equation}
\crossterm (\llambda) =  -\frac{1}{2} \sum_{k=1}^{\k} \kprop \sum_{\br=1}^{\bR} \left[ \m_{\br} \log (2\pi) 
+ \log \det{\Krzzhh}  
 + \postmean{k\br}'\Krzzhhinv \postmean{k\br} + \trace(\Krzzhhinv \postcov{k\br}) \right] .
\label{eq_cross}
\end{equation}
Evaluation of $\crossterm$ involves three potentially costly expressions, $\det{\Krzzhh}$, $\Krzzhhinv \postmean{k\br}$ and $\trace(\Krzzhhinv \postcov{k\br})$, naively $\bigO((\m\Qr)^3)$, considered in turn below. 

\emph{Log determinant:}
Similar to the entropy term, the expression $\det{\Krzzhh}$ decomposes to require only the calculation of $\det{\Krhh}$ and $\det{\Krzz}$ in $\bigO(\m^3 + \Qr^3)$ or $\bigO(\m^3 + \Qr)$ per group for general and sparse models respectively.

\emph{Matrix-vector term:}
%The matrix-vector multiplication $\Krzzhhinv \postmean{k\br}$ can be expanded as 
%
%\begin{equation*}
%\Krzzhhinv \postmean{k\br} = \begin{bmatrix}
%\Krzzinv \left[ \sum_{j=1}^{\Qr} \Krhhinv_{1j} \postmean{k\br j} \right] \\
%\Krzzinv \left[ \sum_{j=1}^{\Qr} \Krhhinv_{2j} \postmean{k\br j} \right] \\
%\vdots \\
%\Krzzinv \left[ \sum_{j=1}^{\Qr} \Krhhinv_{\Qr j} \postmean{k\br j} \right]
%\end{bmatrix}
%\end{equation*}
%
The winged diagonal form of $\Krhhinv$ enables computation of $\Krzzhhinv \postmean{k\br}$ in $\bigO(\m^3 + \Qr)$ in the sparse case versus $\bigO(\m^3 + \Qr^3)$ time in the general case. %We instantiate $\Krhhinv$ using general solvers in $\bigO(\Qr^2)$.

\emph{Trace term:}
The trace term involves both prior and posterior covariance matrices. For the Kronecker posterior which allows 
$\trace(\Krzzhhinv \postcov{k\br}) = \trace(\Krhhinv \postcovkronb)  \trace(\Krzzinv \postcovkronw)$ complexity is reduced from $\bigO(\m^3 + \Qr^3)$ to $\bigO(\Qr + \m^3)$ from the general to sparse.
Where $\postcov{k\br}$ is diagonal, the trace term requires only diagonal elements of $\Krhhinv$, $\Krzzinv$ and $\postcov{k\br}$. 
%Specifically,
%
%$
%\trace(\Krzzhhinv \postcov{k\br}) 
%= \sum_{j=1}^{\Qr} ( \Krhhinv )_{jj} [ \sum_{i=1}^{\m} ( \Krzzinv )_{ii} ( \postcov{k\br j} )_{ii}  ].
%$
%
%By evaluating the diagonal of $\Krhhinv$ in linear time in the sparse case, the cost of the trace term is again reduced 
leading to $\bigO(\Qr + \m^3 + \m\Qr)$ versus $\bigO(\Qr^3 + \m^3 + \m\Qr)$ in the general case. %Again, we instantiate $\Krhhinv$ using general solvers in $\bigO(\Qr^2)$.

%\noteAD{We note here that our experimental implementation of $\Krzzhhinv \postmean{k\br}$ and $\trace(\Krzzhhinv \postcov{k\br})$ currently departs from the achievable linear complexity on sparse matrices by employing some low level linear solve routines in $\bigO(\Qr^2)$. See \S \ref{sec:discussion} for further comments.}
%
\paragraph{Expected Log Likelihood}
$\ellterm$ is defined as
\begin{equation}
\ellterm (\llambda) = \sum_{n=1}^{\n}\expectation{\qnf}{\logpliken}
\label{eq_ell}
\end{equation}
where we estimate the expectation in \eqref{eq_ell} by \gls{MC} sampling from $\qnf$ as discussed below.
\subsection{Indirect sampling}
\label{sec:spggp_indsample}
Sampling from the posterior distribution of $\fn$ and similarly $\fstar$ is required for both $\ellterm$ and prediction. % \citep{Dahl2018} show that 
Samples from $\qnf$ can be drawn from component group posteriors \ie $\qnkfr = \Normal ( \qfmean{k\br (n)}, \qfcov{k\br (n)} )$
 with mean and covariance expressions given by 
$\qfmean{k\br (n)} = \Arn \postmean{k\br}$ and 
$\qfcov{k\br (n)}  = \priorcovrn + \Arn\postcov{k\br}\Arn^{\prime}$, where
\begin{align}
\priorcovrn &= \Krhh \times \left[ \krxxn  \right. \nonumber \\
& \left. - \krxzn \Krzzinv \krzxn   \right] \quad \text{and}  \nonumber \\
\Arn &= \left[ \I_{\Qr} \kron \krxzn \Krzzinv \right]  \nonumber
\end{align}
Given a Kronecker posterior the quadratic term simplifies to
\begin{align*}
&\Arn\postcov{k\br}\Arn^{\prime} = \left[ \postcovkronb \kron \right. \nonumber \\
& \left.  \krxzn \Krzzinv \postcovkronw \Krzzinv \krzxn \right] \nonumber
\end{align*}

Direct sampling as in \citep{Dahl2018} requires factorizing the posterior covariance to obtain a premultiplier $\Psi(\qfcov{k\br (n)})$ such that $\Psi \Psi' = \qfcov{k\br (n)}$.  However, since this differs for each observation the associated cost per group is $\bigO (\n\Qr^3)$, where $\n$ is mini-batch size for $\ellterm$ sampling or $\n_{test}$ for prediction. %Further, batch factorization is inherently unstable.
We avoid this by making use of known Cholesky factors and sampling from two component distributions using the property that the sum of two independent, normally distributed random variables,
$ \X \sim \Normal (\vecmu_{\X}, \Sigma_{\X})$ and $ \Y \sim \Normal (\vecmu_{\Y}, \Sigma_{\Y})$
is also a normally distributed with mean $\vecmu_{\X} + \vecmu_{\Y}$ and covariance $\Sigma_{\X} + \Sigma_{\Y}$.
We draw independent samples from two distributions specified as 
$\Normal(\vec{0}, \priorcovrn)$ and $\Normal(\qfmean{k\br (n)}, \Arn\postcov{k\br}\Arn^{\prime})$, and sum these to return samples from $\Normal ( \qfmean{k\br (n)}, \qfcov{k\br (n)} ) $.  

Complexity of factorizations for $\priorcovrn$ and $\Arn\postcov{k\br}\Arn^{\prime}$ once obtained differs across variants but where a sparse prior is adopted, is reduced to $\bigO(\n\Qr)$.
Critically, in the sparse case memory complexity is significantly reduced by replacing the premultiplier $\Psi(\priorcovrn)$ with vector operations reducing memory complexity from $\bigO(\n\Qr^2)$ to $\bigO(\n\Qr)$.

The time costs to obtain the quadratic term $\Arn\postcov{k\br}\Arn^{\prime}$ are $\bigO(\n\m^2\Qr + \m^3)$, $\bigO(\n(\m^2 + \Qr^2) + \m^3)$ and $\bigO(\n(\m^2 + \Qr) + \m^3)$ for the diagonal, general Kronecker and sparse Kronecker posteriors respectively. Similarly, the time costs of obtaining $\priorcovrn$ are $\bigO(\n(\m^2 + \Qr^2) + \m^3)$ or $\bigO(\n(\m^2 + \Qr) + \m^3)$ for the general and sparse priors respectively. 
Direct sampling would then require combination and factorization of these components in $\bigO(\n\Qr^3)$ time and with $\bigO(\n\Qr^2)$ memory.

Indirect sampling requires factorization of $\priorcovrn$ in $\bigO(\n\Qr^2 + \Qr^3)$ time for general priors 
%\footnote{The total cost of $Chol(\priorcovrn)$ in the general case includes factorisation of $\Krhh$ and the scalar multiplier in $\bigO(\n + \Qr^3)$ plus cross multiplication in $\bigO(\n\Qr^2)$.} 
and $\bigO(\n\Qr)$ for sparse priors. Similarly, given a Kronecker posterior, Cholesky factors for the quadratic term can be obtained in $\bigO(\n\Qr^2 + \Qr^3)$ and $\bigO(\n\Qr)$ for general and sparse specifications, respectively, and $\bigO(\n\Qr)$ in the diagonal case.

%In the sparse case memory complexity is reduced by replacing the premultiplier $\Psi(\priorcovrn)$ (and additionally $\Psi(\Arn\postcov{k\br}\Arn^{\prime})$ for Kronecker posteriors) by vector operations such that premultipliers in $\bigO(\n\Qr^2)$ space are never instantiated. Storage of vector components is reduced to $\bigO(\n\Qr)$ in line with the diagonal premultiplier.

Overall, indirect sampling reduces time, but not memory, complexity relative to direct sampling in the general model. However, by using sparse priors it is possible to achieve significant reductions from the general indirect model. Further gains still are possible if a sparse Kronecker posterior is used in lieu of a diagonal posterior. 
The last point arises since (broadly speaking) operations involving sparse matrices can exploit decomposition properties of the Kronecker product to achieve linear complexity that is additive, rather than multiplicative, over $\m$ and $\Qr$. The same observation may be made about complexity reductions in entropy and cross entropy terms. Further details are provided in the supplement.
\section{Details of experimental set-up}
\label{sec:expts-setup-details}
Three classes of non-\gls{GGP} benchmark models are considered, the linear coregional model (\lcm), \gprn and multi-task model with task-specific features (\mtg) of \citep{Bonilla2008}. These benchmarks are chosen as they can be implemented in the same inference framework as the \gls{GGP} and sparse \gls{GGP},allowing direct comparison of model performance. %\footnote{The multi task model of \citep{Bonilla2008} is not included as prior experiments found the model to be uncompetitive for this problem.} 
For the \lcm model, we report two specifications for $\p=25$. The first mirrors the \gls{GGP} with $\p=\Qg$ %as in \citep{Dahl2018}
-and the second is a lower ranked model with $\Qg=5$ ($\Qg=4$) for diagonal (Kronecker) posteriors. For $\p=50$ we report only the first specification (the best \lcm model based on \rmse). For the \gprn, we report models with $\Qg=2$. This was the best performing specification for $\p=25$, and the largest value for $\Qg$ able to run on the same platform as sparse \gls{GGP} models. 

\subsection{Kernels}
Kernels for \mtg models are defined as $\kernel(f_j(\x), f_{j'}(\xprime) ) = \kernel_{Per.}(t,s) \kernel_{RBF}(\lags_{jt}, \lags_{j's}) \kernel_{RBF-Ep.(2)}(\h_j, \h_{j'})$. 
For \gprn latent functions in a row $\W_i$, the kernel is defined as $\kernel_{Per.}(t,s)\kernel_{RBF}(\lags_{it},\text{ }\lags_{is})$. For node functions, we use a radial basis kernel with lag features. For \lcm $\p=\Qg$, node functions are defined as 
%in \citep{Dahl2018} \ie 
combined periodic-radial basis kernels. For lower rank mixing models, we tested two node specifications for $\Qg = 2,3,4,5$, (a) lags for all sites assigned to each node function, and (b) subsets of lags assigned to node functions based on k-means clustering of lags into $\Qg$ groups. Reported benchmarks for \lcm use clustered lags, while \gprn benchmarks use complete lags per node.

For node functions, $\ksub{\gj}(\x_t,\x_s)$ is a radial basis function kernel on a vector of lagged power features at site $i$, \ie for site $i$ at time $t$, $\lags_{i,t} = (y_{i,t}, y_{i,t-1})$.
For each row-group $r$ in $\Wn$, 
$\krxh = \krxx \krhh$ with
$\krxx = \kernel_{Per.}(t,s)\kernel_{RBF}(\lags_{rt},\lags_{rs})$, where
$ \kernel_{Per.}(t,s)$ is a periodic kernel on a time index $t$.

\subsection{Solar model variants}
\paragraph{Selection of sparsity constraints \apriori}
Given the spatial nature of covariance between sites and where node functions align with tasks, the weight applied to each node by a given task would be expected to be a function of the (notional) spatial location of the node relative to the target task. We assume that weights are conditionally independent given the weight assigned to the task-associated node for that node.
We enforce this constraint for each row-group $i$ of $\Wn$, 
\begin{equation}
\W_{(n)ij} \perp \W_{(n)ik} \quad \lvert \quad \W_{(n)ii}
\label{eq:spggp_CI}
\end{equation}
$j,k \neq i$, $j \neq k$. 
Since latent function groupings in the \gls{GGP} framework can be any subset of $\f$ in any order, it is only required that, for sparse inference, latent functions within each group $\W_{(n)i.}$ are ordered such that $\W_{(n)ii}$ acts as the pivot. 
%Then the Cholesky factor of $\Krhh$ has the form in \eqref{eq_sparsechol_elems}.
%
Within each row of weights, this gives rise to the star configuration for the undirected graphical model (as illustrated at Figure \ref{fig:spggp_ugm} in the supplement), with a single pivot weight function ($\W_{(n)ii}$), and conditionally independent child weight functions corresponding to cross-site nodes ($\W_{(n)ij},j \neq i$).
 
\paragraph{Cross-function kernel variants}
We test three sparse forms of $\Krhh$, which we term (a) `implicitly sparse' (sparsity is automatic \ie implicit in the model due to separable kernel specification as discussed at \S \ref{sec:spggp_implicit_spgp}), (b) `explicitly sparse' (a stationary kernel that does not give rise to automatic sparsity but is used in conjuntion with the constraint in \eqref{eq_cikern} explicitly imposed), and (c) `free sparsity' ($\Krhh$ is freely parameterized using $2\Qr-1$ parameters for nonzero Cholesky elements and no kernel form is defined). 

In the implicitly sparse case, we use a separable, dot product wavelet kernel with Ricker mother wavelet function (see \citep{Zhang2004}) plus diagonal correction as described at \S \ref{sec:spggp_implicit_spgp}, which we assume to be shared across groups. The full kernel is given by 
$\ksub{\br (wav.)} + \ksub{\br (diag)}$ and the (full rank) sparse Cholesky can be constructed as in \eqref{eq_sparsechol_elems} with $\cholK_{ii} = \sqrt{\ksub{\br (diag)}}, i \neq 1$. 

For the explicit sparsity case, we use a combined radial basis, Epanechnikov kernel function %as in \citep{Dahl2018} 
with the constraint at \eqref{eq_cikern} enforced by setting $Chol(\Krhh)_{j,k} = 0$, $j,k \neq 1$, $j \neq k$. 
Specifically, $\krhh = \kernel_{RBF}(\h_j, \h_{j^\prime}) \kernel_{Ep.}(\h_j, \h_{j^\prime})$, $ j,j^\prime=1 \ldots \p$.

\subsection{Experimental Settings}
All models are estimated based on the variational framework explained in \S \ref{sec:inference} with indirect sampling used for both sparse and general models. Starting values for common components were set to be equal across all models and the number of inducing points per group of latent functions is set to approximately standardize computational complexity per iteration relating to $\bR \m^3$, using $\m=200$ per \gls{GGP} group as the baseline.

The \gls{ELBO} is iteratively optimized until its relative change  over successive epochs is less than $10^{-5}$ up to a maximum of 200 epochs or maximum of 20 hours. Optimization is performed using  \adam  \citep{Kingma2014}
with settings $\{LR = 0.005; \beta_{1}=0.09; \beta_{2}=0.99 \}$.  All data except time index features are normalized prior to optimization.
Time and performance measures are captured every five epochs. Timed experimental results reported exclude time required to compute and store the various performance measures during optimization.
%All models are estimated on a multi-GPU machine with four NVIDIA TITAN Xp graphics cards (memory per card 12GB; clock rate 1.58 GHz).  
%Experiments were run until either convergence criteria were reached, or to a maximum of 200 epochs or 1200 minutes runtime.
%
\section{Additional Experimental Results}
\label{sec:additional-expts}
\begin{table}[t]
	\small
	\centering
	\caption{Forecast accuracy and variance of \gls{GGP}, sparse \gls{GGP} and benchmark models. Results shown for diagonal and Kronecker posteriors with $\k=1$ for Adelaide ($\p=25$) and Sydney ($\p=50$) datasets. Results are reported for best performing \gprn and \lcm benchmarks (based on \rmse) and for \lcm where $\Qg=\p$. \mrank is average model rank over accuracy measures (\rmse, \mae and \nlpd).}
	\begin{tabular}{lrrrrr}
		\toprule
		P = 25 &       &       &       &       &  \\
		& \multicolumn{1}{l}{\rmse} & \multicolumn{1}{l}{\mae} & \multicolumn{1}{l}{\nlpd} & \multicolumn{1}{l}{\mrank} & \multicolumn{1}{l}{\fvar} \\
		\midrule
		\textit{Diagonal} &       &       &       &       &  \\
		\gls{GGP}   & 0.343 & 0.216 & 0.399 & 7.3   & 0.121 \\
		\gls{GGP} (free) & 0.345 & 0.215 & 0.378 & 7.0   & 0.116 \\
		Sparse \gls{GGP} (implicit) & 0.344 & 0.217 & 0.423 & 10.7  & 0.124 \\
		Sparse \gls{GGP} (explicit) & 0.345 & 0.215 & 0.382 & 7.7   & 0.115 \\
		Sparse \gls{GGP} (free) & 0.343 & 0.215 & \textbf{0.371} & 4.7   & 0.115 \\
		\lcm ($\Qg=\p$) & 0.371 & 0.235 & 0.553 & 15.0  & 0.181 \\
		\lcm ($\Qg=5$) & 0.366 & 0.239 & 0.481 & 14.3  & 0.151 \\
		\gprn ($\Qg=2$) & 0.344 & 0.217 & 0.451 & 10.7  & 0.149 \\
		\mtg   & 0.381 & 0.241 & 0.501 & 16.7  & 0.171 \\
		&       &       &       &       &  \\
		\textit{Kronecker} &       &       &       &       &  \\
		\gls{GGP}   & 0.343 & 0.213 & 0.382 & 4.3   & 0.117 \\
		\gls{GGP} (free) & 0.344 & \textbf{0.213} & 0.384 & 5.3   & 0.112 \\
		Sparse \gls{GGP} (implicit) & 0.342 & 0.213 & 0.414 & 4.7   & 0.120 \\
		Sparse \gls{GGP} (explicit) & \textbf{0.341} & 0.214 & 0.374 & \textbf{2.7} & 0.114 \\
		Sparse \gls{GGP} (free) & 0.344 & 0.216 & 0.378 & 7.0   & \textbf{0.111} \\
		\lcm ($\Qg=\p$) & 0.375 & 0.236 & 0.583 & 16.0  & 0.180 \\
		\lcm ($\Qg=4$) & 0.367 & 0.240 & 0.475 & 14.7  & 0.147 \\
		\gprn ($\Qg=2$) & 0.342 & 0.214 & 0.446 & 5.7   & 0.150 \\
		\mtg   & 0.381 & 0.237 & 0.502 & 16.3  & 0.170 \\
		&       &       &       &       &  \\
		\midrule
		P = 50 &       &       &       &       &  \\
		\textit{Kronecker} & \multicolumn{1}{l}{\rmse} & \multicolumn{1}{l}{\mae} & \multicolumn{1}{l}{\nlpd} & \multicolumn{1}{l}{\mrank} & \multicolumn{1}{l}{\fvar} \\
		\midrule
		Sparse \gls{GGP} (implicit) & \textbf{0.421} & \textbf{0.254} & \textbf{0.622} & \textbf{1.0} & 0.159 \\
		Sparse \gls{GGP} (explicit) & 0.421 & 0.257 & 0.626 & 2.3   & 0.141 \\
		Sparse \gls{GGP} (free) & 0.423 & 0.258 & 0.625 & 2.7   & \textbf{0.139} \\
		\lcm ($\Qg=\p$) & 0.451 & 0.283 & 0.807 & 5.3   & 0.211 \\
		\gprn ($\Qg=2$) & 0.428 & 0.263 & 0.664 & 4.0   & 0.185 \\
		\mtg   & 0.483 & 0.297 & 0.741 & 5.7   & 0.211 \\
		\bottomrule
	\end{tabular}%
	\label{tab:spggp_results_full}%
\end{table}%

\begin{figure}[htbp]
	\centering
	\includegraphics[width=0.6\linewidth]{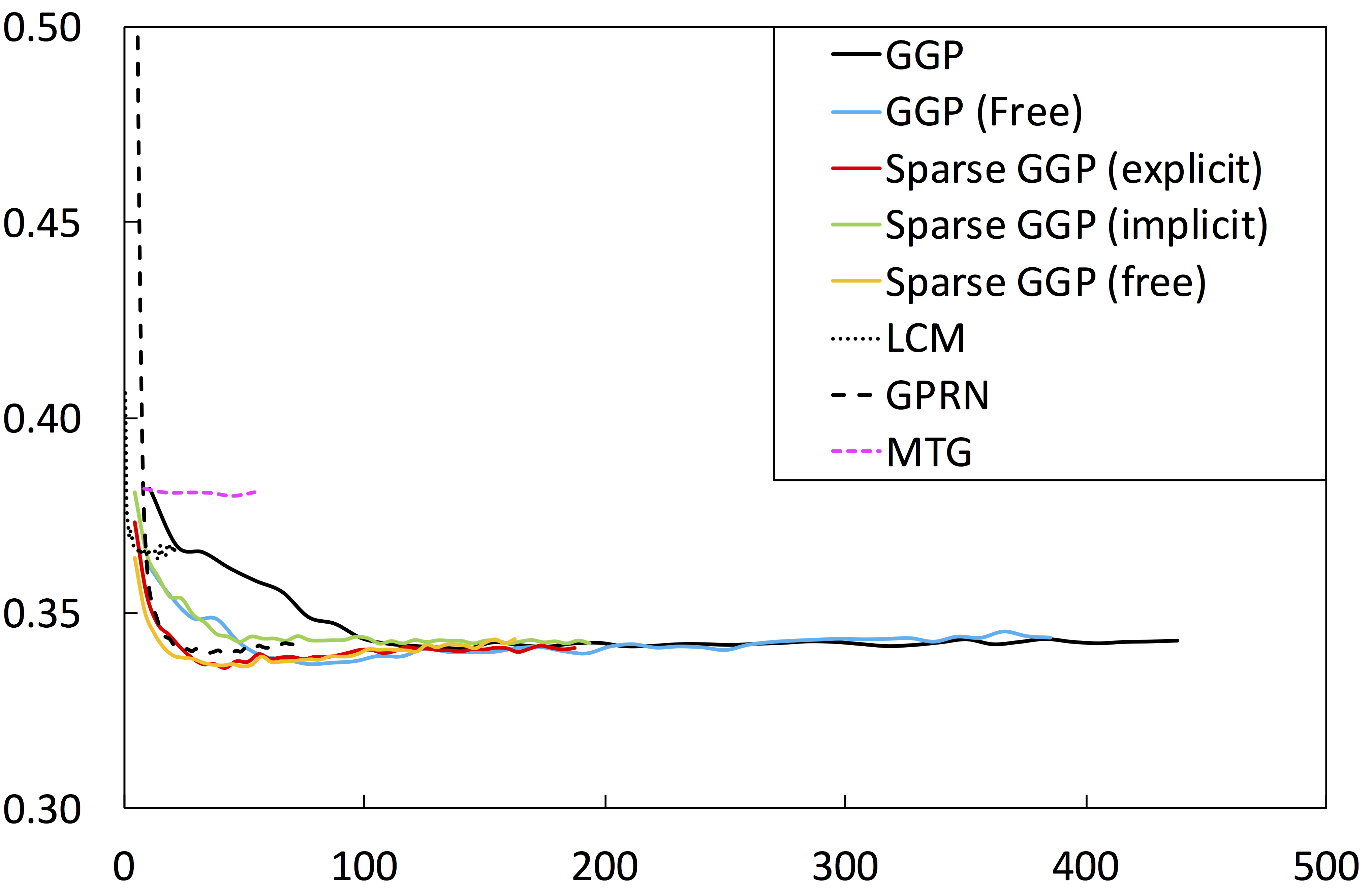}
	\caption{\rmse over optimization time for $\p=25$ with Kronecker posterior.}
	\label{fig:rmsetimep25}
\end{figure}

\begin{figure}[htbp]
	\centering
	\includegraphics[width=0.6\linewidth]{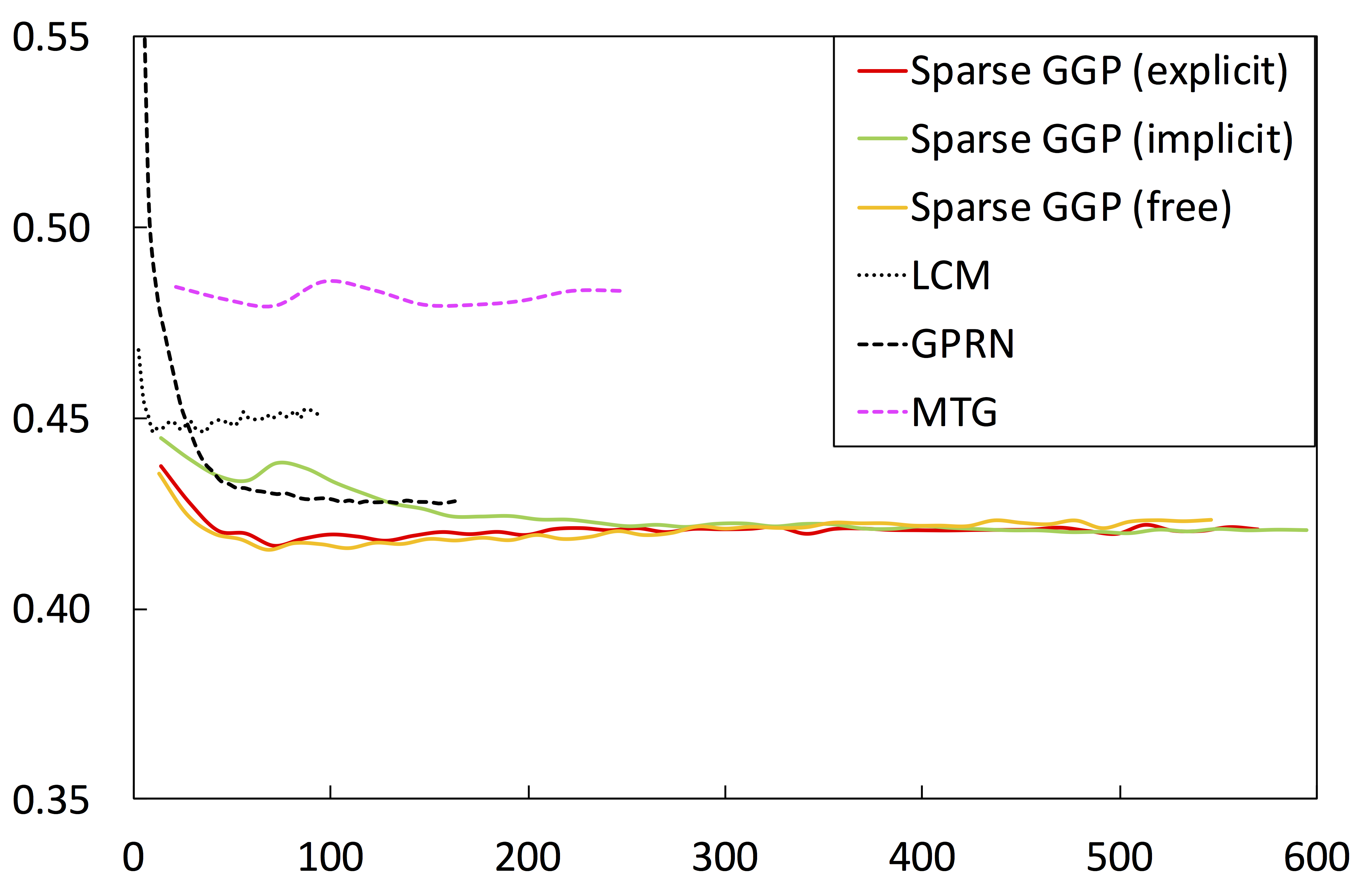}
	\caption{\rmse over optimization time for $\p=50$ with Kronecker posterior.}
	\label{fig:rmsetimep50}
\end{figure}

Further experimental results are provided at Table \ref{tab:spggp_results_full}. Representative results for forecast accuracy over estimation time for \rmse with sparse posterior are shown at Figures \ref{fig:rmsetimep25} and \ref{fig:rmsetimep50}. Optimization of benchmark models is faster than \gls{GGP} based models in some cases, however we also find that performance of \gls{GGP} models quickly surpasses that of benchmark models. We found rankings in terms of speed at which gains are achieved to be consistent across accuracy measures and posterior specifications. In particular, \lcm and \gprn achieve their gains quickly as do explicitly sparse and free sparse \gls{GGP} models, while general \gls{GGP} models and implicitly sparse \gls{GGP} models achieve gains more gradually.

\begin{figure}
	\centering
	\begin{tabular}{cccc}
		\includegraphics[width=0.3\linewidth]{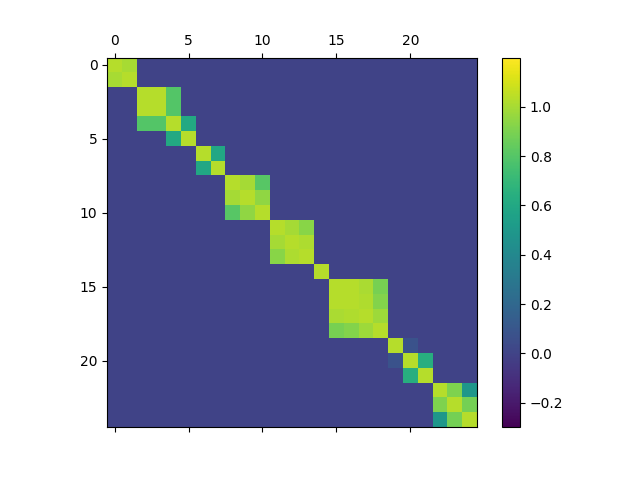}  &  
		\includegraphics[width=0.3\linewidth]{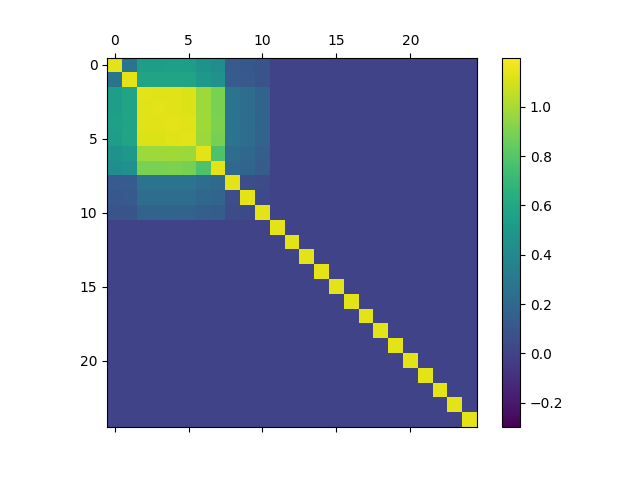} & \includegraphics[width=0.3\linewidth]{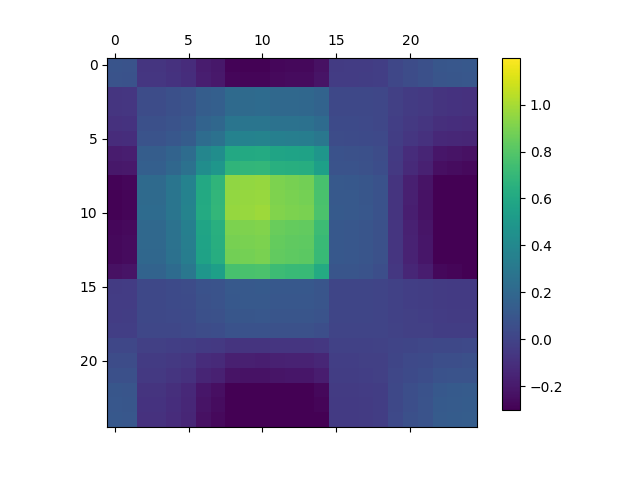} \\

		\includegraphics[width=0.3\linewidth]{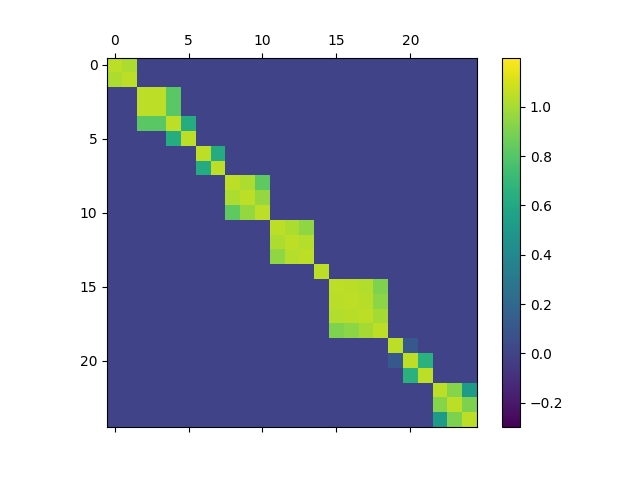}  &  
		\includegraphics[width=0.3\linewidth]{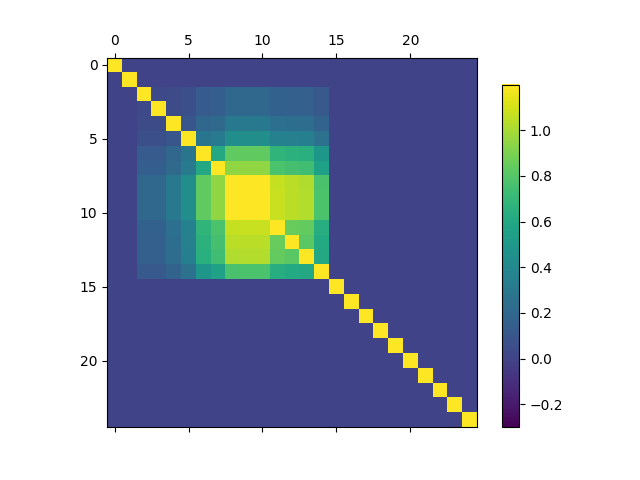} & \includegraphics[width=0.3\linewidth]{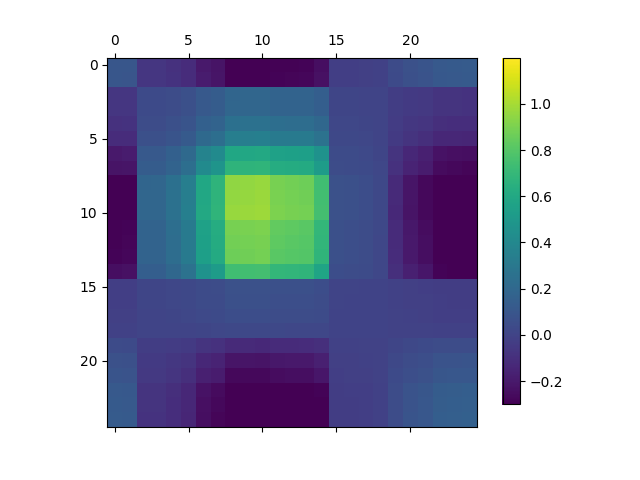} \\

		\includegraphics[width=0.3\linewidth]{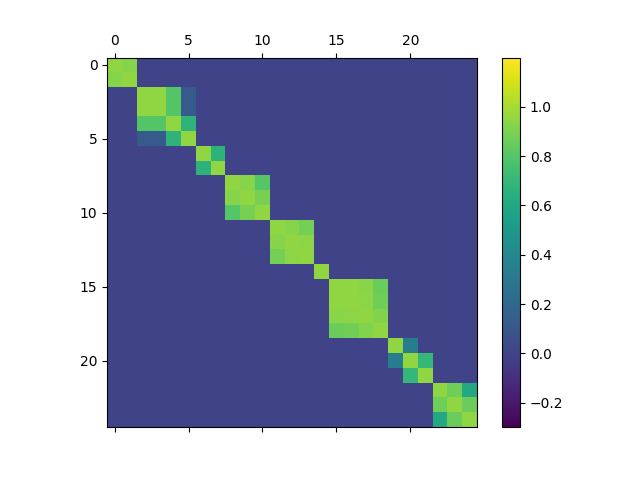}  &  
		\includegraphics[width=0.3\linewidth]{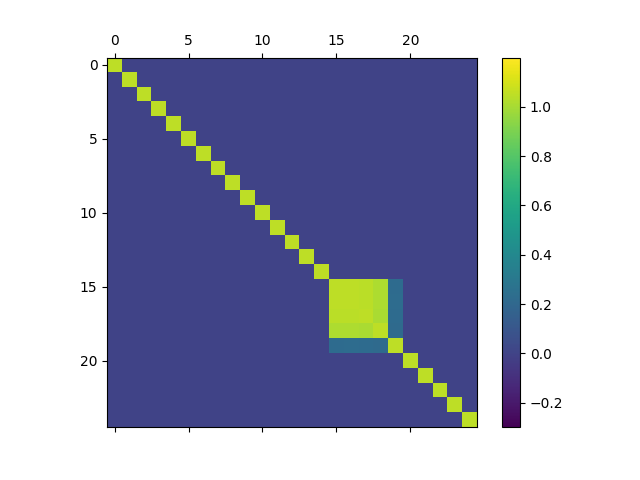} & \includegraphics[width=0.3\linewidth]{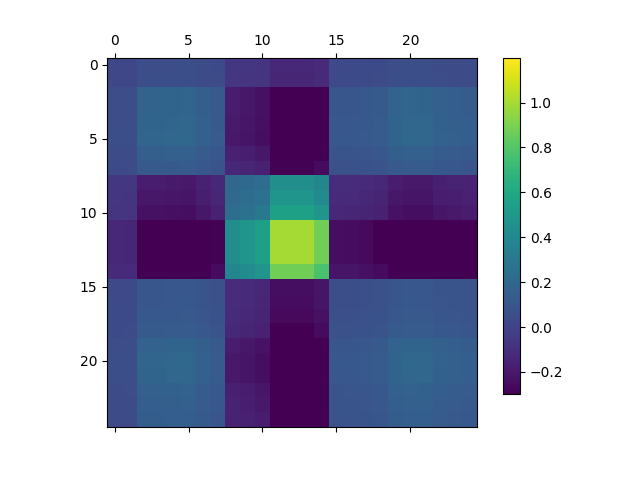} \\

		\includegraphics[width=0.3\linewidth]{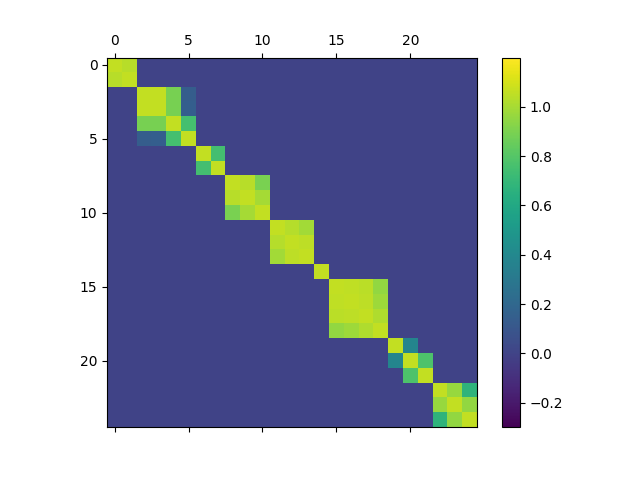}  &  
		\includegraphics[width=0.3\linewidth]{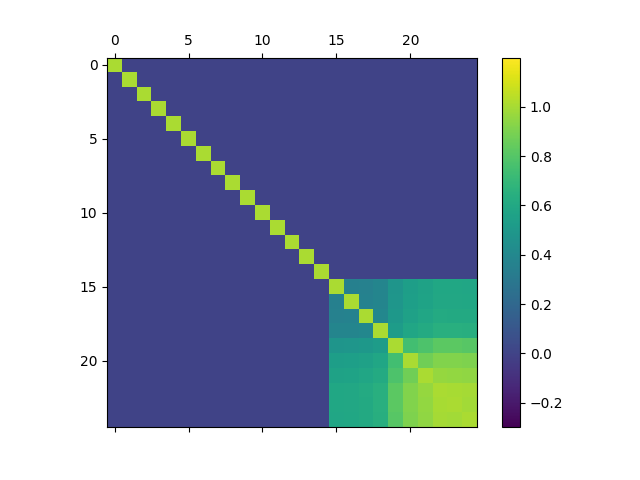} & \includegraphics[width=0.3\linewidth]{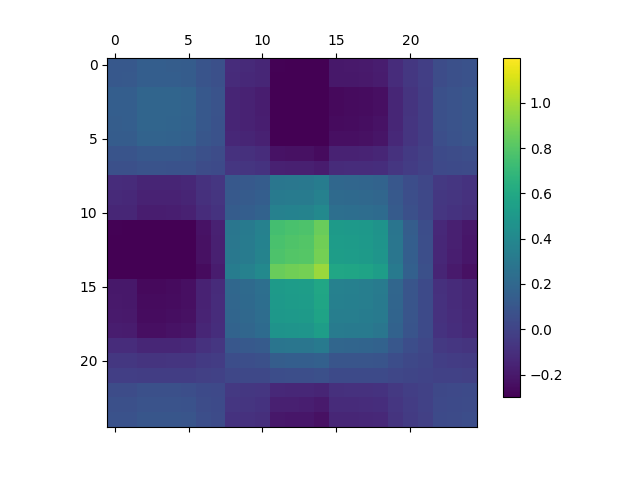} \\

		(a) & (b) & (c) %& (d)
	\end{tabular}
\caption{Examples of $\Krhh$ under \gls{GGP} and sparse \gls{GGP} model variants for $\p=25$ (diagonal). Heatmaps of $\Krhh$ are shown for four row-groups in $\W$ for tasks $i = 4, 10, 16, 22$ for tasks (sites) ordered by site longitude.  Estimated $\Krhh$ shown for \gls{GGP} (Panel (a)), Sparse \gls{GGP} (explicit) (Panel (b)) and Sparse \gls{GGP} (implicit) (Panel (c)). Heatmaps in Panel (c) are shown without the diagonal correction term.}
\label{fig:spggp_heatmaps}
\end{figure}

\paragraph{Estimated covariance}
Illustrative examples of $\Krhh$ estimated under the different \gls{GGP} specifications with diagonal posteriors are presented at Figure \ref{fig:spggp_heatmaps}. Heatmaps are shown for different row groups for sites in longitudinal order. As shown for the initial full \gls{GGP} specification, parameters across row groups were found to be largely consistent with few exceptions (this was also true for latitudinal lengthscales). In contrast, parameters under sparse (explicit) and sparse (implicit) were found to be very adaptive, varying in lengthscale and magnitude. While the explicit sparse construction forces the model to place the most weight on nearest sites, weight centred under the wavelet kernel broadly tended to nearby sites but far less rigidly. 
Freely parameterized models (not shown) did not tend toward covariance structures estimated for models with explicit kernel definitions. Rather, $\Krhh$ in both the full and sparse free models tended to a very sparse diagonal structure.

% panel of heatmap figures for sparse ggp
%

%\bibliography{references}
\bibliography{solar_references,ml_references}
\bibliographystyle{plain}
\end{document}